\documentclass[letterpaper]{article} 
\usepackage{aaai2027}  
\usepackage[hyphens]{url}  
\usepackage{graphicx} 
\usepackage{natbib}  
\usepackage{caption} 
\usepackage{algorithm}
\usepackage{algorithmic}

\usepackage{newfloat}
\usepackage{listings}
\DeclareCaptionStyle{ruled}{labelfont=normalfont,labelsep=colon,strut=off} 
\floatstyle{ruled}
\newfloat{listing}{tb}{lst}{}
\floatname{listing}{Listing}

\usepackage{booktabs}

\usepackage{amsfonts}
\usepackage{booktabs}
\usepackage{multirow}
\usepackage{graphicx}
\graphicspath{{img/}}
\usepackage{amsmath}
\usepackage{subcaption}
\usepackage{tabularx}

\title{Contrastive Mixed Prompt Learning for Incomplete Multimodal Sentiment Analysis with Unseen Modality Combination}
\author{
    Kaixin Xu\textsuperscript{\rm 1},
    NaiJin Liu\textsuperscript{\rm 2},
    Yulin Kang\textsuperscript{\rm 3},
    Tangyue Jin\textsuperscript{\rm 4},
    Zixuan Yu\textsuperscript{\rm 5},
    Wenxi Zhao\textsuperscript{\rm 5},
    Yibei Liu\textsuperscript{\rm 5},
    Qianle Zhang\textsuperscript{\rm 6},
    Yangyang Wu\textsuperscript{\rm 1}\corresponding,
    Mengying Zhu\textsuperscript{\rm 1},
    Meng Xi\textsuperscript{\rm 1}
}
\affiliations{
    \textsuperscript{\rm 1}School of Software Technology, Zhejiang University\hspace{1em}
    \textsuperscript{\rm 2}Beijing WuZi University\\
    \textsuperscript{\rm 3}Ant Group \hspace{1em}
    \textsuperscript{\rm 4}China University of Geosciences\\
    \textsuperscript{\rm 5}University of Electronic Science and Technology of China\hspace{1em}
    \textsuperscript{\rm 6}South China University of Technology
}

\nocopyright
\begin{document}

\maketitle

\begin{abstract}
Incomplete multimodal sentiment analysis has garnered significant attention in recent years.
Existing approaches typically assume that data is missing at random or are designed specifically for certain missing patterns, ignoring the modality combination inconsistency between training and testing phases. 
However, in real-world scenarios, the testing phase often encounters modal combinations that were not present during the training phase, which leads to insufficient generalization capabilities and unstable performance. 
In this paper, we introduce the problem of Incomplete Multimodal Sentiment Analysis with Unseen Modality Combinations (IMSAUMC), aiming to enhance model generalization for unseen modality combinations.
To address this challenge, we propose the model named \textbf{C}ontrastive \textbf{M}ixed \textbf{P}rompt \textbf{L}earning (\textsf{CMPL}) for IMSAUMC.
It introduces a label-guided contrastive feature learning mechanism to learn robust and discriminative cross-modal representations.
Additionally, we design modality-combination prompts with a soft router to facilitate better learning of various modality combinations.
Furthermore, we introduce three prompt contrastive learning strategies, which enable effective learning of prompts corresponding to unseen modality combinations, thereby significantly strengthening the model's generalization capabilities in diverse testing scenarios.
Extensive experiments on three widely used datasets demonstrate that \textsf{CMPL} achieves more than a 5\% improvement in accuracy compared to state-of-the-art approaches.

\end{abstract}


\section{Introduction}
\label{sec:intro}

Multimodal Sentiment Analysis (MSA) has emerged as a pivotal research area in recent years. By jointly modeling textual utterances, acoustic characteristics, and visual expressions, MSA systems aim to infer human affective states with significantly enhanced robustness compared to unimodal approaches.
Many works~\cite{hu2018multimodal,zhu2022multimodal,huang2024tmbl,liu2024ensemble, wang2025dlf} have achieved promising results by exploiting cross-modal complementarity.

However, real-world deployment scenarios frequently violate the full-modality assumption.
Practical challenges such as sensor failures, background noise, occlusions, and privacy constraints often lead to missing modalities during inference.
Consequently, substantial research efforts have been directed toward MSA with missing modality~\cite{pham2019found, ma2021smil, huan2023unimf, li2024unified, li2024toward}.

\begin{figure}[t]
  \centering
  \includegraphics[width=1\linewidth]{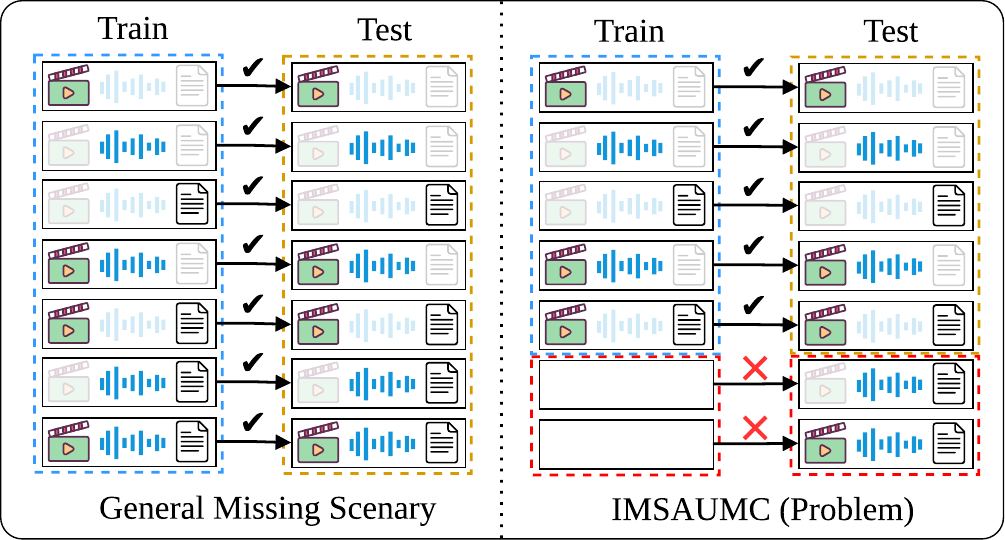}
  \caption{The difference between the general MSA with missing modality task and the IMSAUMC task.
  In the general MSA with missing modality task, all possible modality combinations are available during both the training and testing phases.
  However, in our IMSAUMC task, only partial modality combinations are present during training, while unseen modality combinations may appear during testing.}
  \label{fig:task}
\end{figure}

Although these approaches demonstrate resilience against random modality absence, they fundamentally assume that all modality combinations have been observed during training.
In real-world data, however, missing patterns are often structured rather than random.
For example, a camera failure causes the absence of visual features; a subsequent audio failure then results in missing the visual-audio modality combination entirely.
Consequently, the dataset may contain only a subset of possible combinations, and the test set is likely to encounter unseen ones.
For another example, users on social media may decline to upload certain modalities (e.g., audio) due to privacy concerns, yielding datasets with only partial combinations.
In such scenarios, we aim to train models using only the available partial combinations, while still ensuring strong generalization when more complete modality combinations appear at inference time.

For a dataset with $n$ modalities, $2^n-1$ possible combinations exist. As illustrated above, the dataset may not cover all missing patterns, and the test set may encounter unseen modality combinations, especially as $n$ grows. Consequently, existing methods often struggle to handle such scenarios effectively, and approaches capable of generalizing to unseen modality combinations are critically needed.


We introduce the task of \textbf{I}ncomplete \textbf{M}ultimodal \textbf{S}entiment \textbf{A}nalysis with \textbf{U}nseen \textbf{M}odality \textbf{C}ombinations (IMSAUMC), which handles missing modalities and unseen modality combinations during testing, as shown in Figure~\ref{fig:task}.
While previous work in MSA with missing modalities has achieved notable progress, two key challenges remain:
First, many works employ contrastive learning to obtain modality-invariant representations. However, most fail to account for the intrinsic data structure, potentially separating semantically similar samples, which leads to suboptimal representations (\textbf{CH1}).
Second, in scenarios involving unseen modality combinations, existing methods often overlook the relations between different modality combinations.
As a result struggle to effectively handle test-time data whose modality combinations were not encountered during the training phase (\textbf{CH2}).

To address these challenges, we propose \textbf{C}ontrastive \textbf{M}ixed \textbf{P}rompt \textbf{L}earning (\textsf{CMPL}) for multimodal sentiment analysis with unseen modality combinations.
Specifically, for \textbf{CH1}, we introduce label-guided contrastive feature learning, which incorporates label similarity constraints to pull representations of same-labeled samples closer while maintaining distances proportional to label dissimilarity for others.
For \textbf{CH2}, we design mixed modality-combination prompts with a soft routing mechanism that dynamically selects prompts to comprehensively model inter-combinatorial relations.
Moreover, we develop three prompt contrastive learning strategies to further enhance generalization to unseen combinations: a modality information preservation strategy, a cross-combination complementarity strategy, and a conditional information alignment strategy.
The main contributions are summarized as follows:
\begin{itemize}
    \item We propose \textsf{CMPL}, a novel model for the IMSAUMC task that improves generalization to unseen modality combinations. To the best of our knowledge, this is the first work addressing this problem.
    \item We propose a \emph{label-guided contrastive feature learning} mechanism, which enforces representation consistency for samples with identical labels while constraining the distance between dissimilar samples proportionally to their label differences. This preserves the structural relationships within sample representations.
    \item We introduce a \emph{mixed prompts learning} mechanism coupled with three \emph{prompt contrastive learning} strategies. These comprehensively model inter-combinatorial relationships and enhance the model's generalization to unseen modality combinations.
    \item Extensive experiments on CMU-MOSI, CMU-MOSEI, and SIMS-V2 datasets demonstrate the effectiveness of our method over state-of-the-art approaches.

\end{itemize}

\section{Related Work}
\label{sec:related}

\subsection{Multimodal Sentiment Analysis}
Multimodal Sentiment Analysis (MSA) aims to infer sentiment by integrating heterogeneous data from multiple modalities, such as text, visual, and acoustic signals.
MSA methods~\cite{truong2019vistanet, yu2021learning, mai2022hybrid, sun2022cubemlp, li2023decoupled} leverage cross-modal complementarity to improve robustness and accuracy over unimodal approaches.
For instance, \citeauthor{yu2021learning} proposed self-MM~\cite{yu2021learning}, jointly training a multimodal main task with unimodal subtasks as pseudo-label supervision to learn inter-modal consistency and cross-modal differences.
\citeauthor{sun2022cubemlp} presented Cube-MLP~\cite{sun2022cubemlp}, which mixes features along three axes via MLP units.
\citeauthor{li2023decoupled} introduced DMD~\cite{li2023decoupled}, decoupling homogeneous and heterogeneous features with adaptive cross-modal distillation to enhance modality discriminability.
\citeauthor{wang2025dlf} proposed DLF~\cite{wang2025dlf}, a disentangled language-focused framework that reduces cross-modal redundancy for improved MSA performance.

\begin{figure*}[t]
  \centering
  \includegraphics[width=\linewidth]{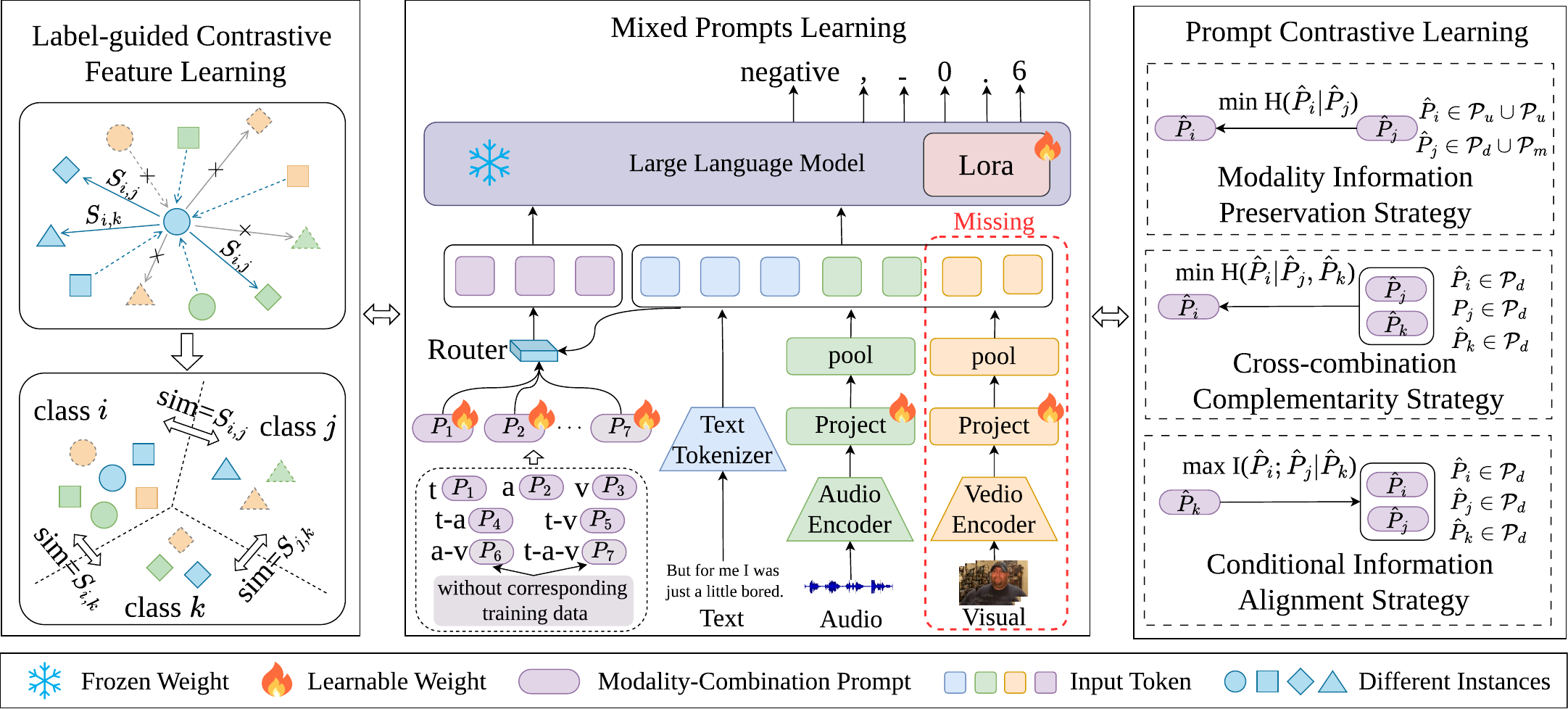}
  \caption{The framework of our \textsf{CMPL}, which consists of three components: the \emph{labeled-guided contrastive feature learning} module, the \emph{mixed prompts learning} mechanism, and \emph{prompt contrastive learning} strategies. 
  Taking the absence of audio-visual and text-audio-visual modality combinations in training data, as well as the missing vision modality in input, as an example.
}
  \label{fig:model}
  \vspace*{-0.1in}
\end{figure*}

However, real-world data often suffers from missing modalities. 
Many methods~\cite{yuan2021transformer, zeng2022tag, yuan2023noise, xu2024leveraging, zhang2024towards} have been developed to address MSA with missing modality.
For example, TFR-Net~\cite{yuan2021transformer} employs a feature reconstruction module to generate missing modality content.
LNLN~\cite{zhang2024towards} improves robustness by guaranteeing high-quality dominant modality representation.
HME~\cite{zhuanghyper} leverages cross-sample semantic enrichment and uncertainty-aware fusion, eliminating explicit modality reconstruction while enhancing robustness and generalization.
MFMB-Net~\cite{tao2025multi} jointly performs global–local dual-stream fusion and collaborative feature reconstruction to robustly handle missing modalities.
However, these methods overlook inconsistent distributions of modality combinations in missing-modality scenarios, where the test set may contain unseen modality combinations that were not present during training.
In contrast, our approach focuses on leveraging known modality combination information to enhance generalization to unseen combinations.


\subsection{Prompt Learning}
Prompt learning has emerged as a powerful paradigm for adapting pre-trained models such as large language models (LLM) to downstream tasks~\cite{gao2021making, heinzerling2021language, liang2022modular, zhu2023prompt}.
\citeauthor{tsimpoukelli2021multimodal} concatenated visual embeddings as prefix prompts to enable frozen language models to generate appropriate captions~\cite{tsimpoukelli2021multimodal}.
\citeauthor{lee2023multimodal} designed missing-aware prompts for different missing-modality cases to enhance robustness~\cite{lee2023multimodal}.
\citeauthor{khattak2023maple} designed branch-aware multi-modal prompts to enhance alignment between language and visual modalities~\cite{khattak2023maple}.
MPLMM~\cite{guo2024multimodal} generates missing modality features and strengthens intra- and inter-modality learning by designing generative, missing-signal, and missing-type prompts.
These methods ignore the relations between modality combinations, limiting generalization to unseen combinations. 
In contrast, our approach thoroughly explores inter-modal relationships and designs three prompt contrastive learning strategies to address unseen modality combinations.

\section{Methodology}

\subsection{Problem Formulation}
Given a multimodal dataset $\mathbf{X} = \{\mathbf{X}^t, \mathbf{X}^a, \mathbf{X}^v\}$ with three modalities (text, audio, visual), each $\mathbf{X}^k = \{{\mathbf{x}^k_1, \cdots, \mathbf{x}^k_N}\} \in \mathbb{R}^{N \times l_k \times d_k}$ denotes the feature matrix of modality $k$, where $N$ is the number of samples, and $l_k$, $d_k$ are the sequence length and embedding dimension, with $k \in \{t, a, v\}$.
The sentiment labels are $\mathbf{Y} \in \mathbb{R}^N$.
For missing modalities, we define mask matrices $\mathbf{M}^t$, $\mathbf{M}^a$, $\mathbf{M}^v$ where $\mathbf{M}^k_i = 0$ indicates the $i$-th sample is missing in modality $k$ and $\mathbf{M}^k_i = 1$ indicates its presence.
The textual features are obtained from text embeddings of a LLM, while audio and visual features are extracted using pre-trained toolkits.
For these three modalities, excluding the all-absent case, there are $T=7$ possible modality combinations, denoted as $\mathcal{S} = \{S_1, S_2, \cdots, S_T\}=\{\{t\},\{a\},\{v\},\{t,a\},\{t,v\},\{a,v\},\{t,a,v\}\}$.
For example, $S_4$ represents the text-audio combination with visual modality missing, with available data $\mathbf{x}=\{\mathbf{x}^t, \mathbf{x}^a\}$.
More detailed information can be found in Appendix \ref{sec:task_desc}.

\emph{Definition.}
The objective of IMSAUMC is to train a model for sentiment analysis under the condition that the training data only contains a subset of the modality combinations in $\mathcal{S}$, while the test data includes all possible modality combinations (\textit{i.e.}, $\mathcal{S}$). 
The model must generalize to unseen modality combinations during testing.

\subsection{Overall Framework}
Figure~\ref{fig:model} shows the framework of our \textsf{CMPL}.
First, during the representation learning stage, we extract sequential embeddings for the audio and visual modalities using pre-trained tools and reduce the sequence length via adaptive average pooling. 
For the textual modality, we leverage text embeddings from the LLM to obtain contextually correlated embeddings.
Then, guided by label similarity, we apply contrastive learning across modality embeddings to learn robust representations while preserving semantic structure.
Additionally, we equip each modality combination with a prompt. 
The multimodal embeddings are then fed into a router to generate a mixed prompt.
Furthermore, three prompt contrastive learning strategies are employed to exploit relations between modality combinations, improving generalization to unseen modality combinations.
Finally, we train \textsf{CMPL} with the following objective function:
\begin{equation}
\mathcal{L}_\textsf{CMPL}=\mathcal{L}_{task}+\alpha\cdot\mathcal{L}_{lcfl}+\beta\cdot\mathcal{L}_{pcl},
\end{equation}
where $\mathcal{L}_{task}$, $\mathcal{L}_{lcfl}$, and $\mathcal{L}_{pcl}$ are task-specific loss, label-guided contrastive loss, and prompt contrastive loss, respectively.
Here, $\mathcal{L}_{task}$ is used to guide the model's predictions, and we employ the traditional autoregressive cross-entropy loss from LLMs for this purpose.
The parameters $\alpha$ and $\beta$ are the balanced factors on $\mathcal{L}_{lcfl}$ and $\mathcal{L}_{pcl}$, respectively.

\subsection{Label-guided Contrastive Feature Learning}
Contrastive learning, as an effective representation learning method, has been widely applied in multimodal domains.
Existing methods typically maximize the similarity of representations across multiple modalities while minimizing the similarity between different samples directly. 
However, these approaches often overlook the structural relationships between samples, potentially separating representations of similarly labeled samples and leading to suboptimal representations.

To address these challenges, we propose the \emph{label-guided contrastive feature learning} (LCFL) module.
This mechanism aims to pull together latent representations of different modalities from the same class while preserving similarity between samples of related classes. 
By incorporating class-aware semantic relationships, this mechanism enables the learned representations to incorporate richer and more robust semantic information, enhancing the model's understanding of multimodal representations.

Specifically, for audio and vision modalities, we first project the available embeddings into the LLM's text embedding space and use an adaptive pooling to reduce the sequence~\cite{yao2024deco} as follows:
\begin{equation}
\begin{aligned}
    \bar{\mathbf{X}}^a &= \text{Pool}(\sigma(\mathbf{X}^a\cdot\mathbf{W}^a_1)\cdot\mathbf{W}^a_2), \\
\end{aligned}
\end{equation}
where $\sigma$ is the activation function.
$\mathbf{W}^a_1$ and $\mathbf{W}^a_2$ are trainable parameters.
The formulation for $\bar{\mathbf{X}}^v$ follows analogously.
$\bar{\mathbf{X}}^a \in \mathbb{R}^{N \times l \times d}$ and $\bar{\mathbf{X}}^v \in \mathbb{R}^{N \times l \times d}$ are the audio and visual embeddings after projection and adaptive pooling, respectively.
Then, we project the embeddings of all modalities into the contrastive learning space~\cite{chen2020simple} as follows:
\begin{equation}
    \begin{aligned}
        \mathbf{H}^{k} = \sigma(\mathbf{X}^{k}\cdot\mathbf{W}^{k}_1)\cdot\mathbf{W}^{k}_2, 
    \end{aligned}
\end{equation}
where ${k} \in \{t,a,v\}$.
$\mathbf{H}^{k} \in \mathbb{R}^{N \times l \times d_c}$ is the representation of modality $k$ after projection.
For representation $\mathbf{H}^w_i$, we treat the instances have the same label as positive pairs, denoted as $\mathbf{H}^u_j|_{\mathbf{Y}_j=\mathbf{Y}_i,\ u\neq w |j\neq i}$, while considering others as negative pairs, denoted as $\mathbf{H}^u_j|_{\mathbf{Y}_j \neq \mathbf{Y}_i}$, where $u, w\in\{t, a, v\}$.
We use the cosine distance to evaluate the similarity between $\mathbf{H}^{w}_i$ and $\mathbf{H}^{u}_j$:
$d(\mathbf{H}^{w}_i,\mathbf{H}^{u}_j)={\langle \mathbf{H}^{w}_i,\mathbf{H}^{u}_j \rangle}/{\Vert \mathbf{H}^{w}_i \Vert \cdot \Vert\mathbf{H}^{u}_j \Vert},$
where $\langle \cdot, \cdot\rangle$ is the dot product operator.

To effectively enhance the model's comprehension of multimodal embeddings and explore cross-modal relationships, we design a LCFL loss function $\mathcal{L}_{lcfl}$.
Our method maximizes similarity between positive pairs while maintaining the similarity of negative pairs according to their label relations. 
This approach effectively mitigates the adverse effects of incorrectly pushing apart embeddings sharing similar labels.
Given text and audio modality as an example, the contrastive loss $\mathcal{L}^{(t,a)}$ between $\mathbf{H}^{t}$ and $\mathbf{H}^{a}$ can be defined as:
\begin{equation}\label{eq:L_t_a}
        \mathcal{L}^{(t,a)} = -\frac{1}{2N}\sum_{w=t,a}\sum_{i=1}^{N}\mathbf{M}^w_i\log\frac{\mathcal{T}^w_{i}}{\mathcal{T}^w_{i}+\mathcal{N}^w_{i}},
\end{equation}
where {\small$
        \mathcal{N}^{w}_{i} =\sum_{j=1}^{N}\sum_{u=t,a}{\mathbf{M}^u_j\cdot\mathbb{I}_{[S_{i,j}\neq 1]}\cdot e^{|d(\mathbf{H}^{w}_i,\mathbf{H}^{u}_j)-S_{i,j}|/\tau}}$}
and {\small$
        \mathcal{T}^{w}_{i} = \sum_{j=1}^{N}\sum_{u=t,a}{ \mathbf{M}^u_j\cdot\mathbb{I}_{[S_{i,j}=1]} \cdot e^{d(\mathbf{H}^{w}_i,\mathbf{H}^{u}_j)/\tau}}-e^{1/\tau}$}.
$\tau$ is the temperature parameter that adjusts the
softness. 
$N$ represents the number of instances.
$\mathbb{I}_{[S_{i,j}=1]}$ is the indicator function that equals 1 \emph{iff} $S_{i,j}=1$.
$S_{i,j}$ is the similarity between labels of $i$-th and $j$-th instance.
Here, we employ a simple approach to measure inter-label similarity:
\begin{equation}
    S_{i,j}=1-\frac{\left|\mathbf{Y}_j-\mathbf{Y}_i\right|}{\max\{\mathbf{Y}\} - \min \{\mathbf{Y}\} },
\end{equation}
where $\max\{\mathbf{Y}\}$ and $\min \{\mathbf{Y}\} $ represent the maximum and minimum values of the labels, respectively.

Similarly, we can compute $\mathcal{L}^{(t,v)}$ and $\mathcal{L}^{(v,a)}$.
Then, the objective function $\mathcal{L}_{lcfl}$ can be calculated as follows:
\begin{equation}
    \mathcal{L}_{lcfl}=\mathcal{L}^{(t,a)}+\mathcal{L}^{(t,v)}+\mathcal{L}^{(v,a)}.
\end{equation}


By minimizing $\mathcal{L}_{lcfl}$, the representations of instances with consistent labels are pulled closer, while the similarity between others aligns with their label similarity. 
This captures more structured representations and effectively enhances multimodal learning in incomplete combinations, effectively facilitating learning for unseen modality combinations.

\subsection{Mixed Prompts Learning}
Most LLM fine-tuning methods employ LoRA for adaptation. 
However, in the IMSAUMC task, using LoRA alone to simultaneously train multiple modality combinations fails to distinguish between them.
To address this, we design a \emph{mixed prompts learning} mechanism to enhance the model's understanding of diverse modality combinations.

Specifically, we design \emph{modality-combination prompts}, \textit{i.e.}, $\mathcal{P} = \{\mathbf{P}^1, \mathbf{P}^2, \cdots, \mathbf{P}^T\}$, where $\mathbf{P}^i\in \mathbb{R}^{l_p \times d_p}$ is the prompt embedding for the $i$-th modality combination, with $l_p$ and $d_p$ being its sequence length and dimension.
Traditional methods concatenate each modality-combination prompt with its corresponding input and feed them into the LLM, learning the prompts from available data.
However, in IMSAUMC tasks, the training set does not cover all modality combinations appearing in the test set. Consequently, prompts for unseen combinations lack training data and cannot be learned.
Recognizing that different modality combinations are not independent but exhibit correlations, we propose a \emph{Soft Routing-inspired mixed prompts guidance} mechanism.
Given multimodal data $\mathbf{X}_j=[\mathbf{X}_j^t,\bar{\mathbf{X}}_j^a] \in \mathbb{R}^{N \times 2l \times d}$, where $[\cdots]$ denotes sequence concatenation, this mechanism feeds $\mathbf{X}_j$ into a router that automatically selects and weights prompts according to the input representation:
\begin{equation}
    G(\mathbf{X}_j)=\text{Softmax}(\mathbf{X}_j\cdot \mathbf{W}_g),
\end{equation}
where $\mathbf{W}_g \in \mathbb{R}^{d \times T}$ is the router's parameter and $G(\mathbf{X}_j)$ represents the soft assignment weights.
Subsequently, we obtain the final mixed prompt via dynamic blending:
\begin{equation}
    \bar{\mathbf{P}}_j=\sum\nolimits_{i=1}^{T}{G(\mathbf{X}_j)_i\cdot\mathbf{P}^i},
\end{equation}
where $\bar{\mathbf{P}}_j$ is the mixed prompt for the $j$-th instance.

Finally, the mixed prompt $\bar{\mathbf{P}}_j$ and multimodal input $\mathbf{X}_j$ are jointly fed into the LLM to produce the output:
\begin{equation}
    \bar{\mathbf{Y}}_j=\text{LLM}(\bar{\mathbf{P}}_j,\mathbf{X}_j;\theta),
\end{equation}
where $\theta$ represents the LLM's parameters and $\bar{\mathbf{Y}}_j$ is the generated text with sentiment class and sentiment score.
Following standard LLM training, we adopt next-token prediction loss. Thus, the task loss $\mathcal{L}_{task}$ is:
\begin{equation}
    \mathcal{L}_{task}=\sum\nolimits_{i=1}^{N}\sum\nolimits_{j=1}^{K}{-\log P(L_{i,j}|\bar{\mathbf{P}_i},\mathbf{X}_i,\theta)},
\end{equation}
where $K$ is the number of label tokens and $L_{i,j}$ is the $j$-th label token of the $\bar{\mathbf{Y}}_i$ generated by the $i$-th instance.

\subsection{Prompt Contrastive Learning}
In the IMSAUMC task, a key challenge lies in effectively leveraging knowledge from existing modality combinations to enhance the learning of unseen modality combinations. 
Recognizing that different modality combinations are not isolated but inherently interrelated, we design three prompt contrastive learning strategies to enable mutual learning among prompts:
(1) \emph{modality information preservation} strategy;
(2) \emph{cross-combination complementarity} strategy;
(3) \emph{conditional information alignment} strategy.
The core idea is to consider the relationships between various modality combinations to minimize the conditional entropy or maximize the conditional mutual information.

To compute conditional entropy and mutual information, we first project the modality-combination prompt embeddings, then average them along the sequence dimension, and finally apply the softmax function, which allows the prompt representation to be interpreted as a probability distribution, enabling entropy and mutual information estimation:
\begin{equation}\label{eq:P_hat}
\hat{\mathbf{P}}^i=\text{Softmax}(\text{Avg}(\sigma(\mathbf{P}^i\cdot\mathbf{W}^i_1)\cdot\mathbf{W}^i_2)),
\end{equation}
where $\hat{\mathbf{P}}^i \in \mathbb{R}^{l'_p\times D}$ are the normalized prompt embeddings of $i$-th modality combination.
For convenience, we denote the set of unimodal prompts as $\mathcal{P}_u=\{\hat{\mathbf{P}}^1, \hat{\mathbf{P}}^2, \hat{\mathbf{P}}^3\}$, the set of dual modality prompts as $\mathcal{P}_d=\{\hat{\mathbf{P}}^4, \hat{\mathbf{P}}^5, \hat{\mathbf{P}}^6\}$, and the set of full modality prompts as $\mathcal{P}_f=\{\hat{\mathbf{P}}^7\}$.

\emph{Modality Information Preservation.}
For a multimodal prompt, it inherently contains the information present in each of its unimodal components. 
Therefore, it can be argued that when given a multimodal prompt, it should retain the sub-modality-combination prompts it encompasses as much as possible. 
To achieve this, we minimize the conditional entropy $H(\hat{\mathbf{P}}^i \mid \hat{\mathbf{P}}^j)$ between such sub-modality-combination $\hat{\mathbf{P}}^i$ and multi-modality-combination prompt $\hat{\mathbf{P}}^j$.
Since each element of $\hat{\mathbf{P}}^i$ and $\hat{\mathbf{P}}^j$ can be treated as probability distribution of two variables $z_i$ and $z_j$ over $D$ classes~\cite{ji2019invariant,huang2020deep,lin2021completer}, where $D$ is the dimensionality of $\hat{\mathbf{P}}^i$ and $\hat{\mathbf{P}}^j$.
The joint probability distribution $P^{(m,n)}\in \mathbb{R}^{D\times D}$ can be defined as:
\begin{equation}\nonumber
    P^{(m,n)}_{i,j}=\frac{1}{l}\sum_{k=1}^{l}{\hat{\mathbf{P}}^m_{k,i}\hat{\mathbf{P}}^n_{k,j}}.
\end{equation}
Let $P^{(m,n)}_d$ and $P^{(m,n)}_{d'}$ denote the margin probability distributions $P^{(m,n)}(z_m=d)$ and $P^{(m,n)}(z_n=d')$, which can be obtained
by summing the $d$-th rows and $d'$-th columns of $P$.
We can define the loss function between the $m$-th prompt and the $n$-th prompt $\mathcal{L}^{(m,n)}$ as follows:
\begin{equation}\nonumber
    \mathcal{L}^{(m,n)}=H(\hat{\mathbf{P}}^m \mid \hat{\mathbf{P}}^n)=-\sum_{d=1}^{D}\sum_{d'=1}^{D}{}P^{(m,n)}_{d,d'}\ln\frac{P^{(m,n)}_{d,d'}}{P^{(m,n)}_{d'}}.
\end{equation}
The loss function $\mathcal{L}_{mip}$ can be defined as follows:
\small{
\begin{equation}\nonumber
\mathcal{L}_{mip}=\sum_{m=1,2}\mathcal{L}^{(m,4)}+\sum_{m=1,3}\mathcal{L}^{(m,5)}+\sum_{m=2,3}\mathcal{L}^{(m,6)}+\sum_{m=1}^6\mathcal{L}^{(m,7)}.
\end{equation}
}

\emph{Cross-Combination Complementarity}.
For two prompts that share a common modality—such as $\hat{\mathbf{P}}^4$ (text-audio combination) and $\hat{\mathbf{P}}^5$ (text-visual combination), their shared textual information acts as a bridge connecting the other two modalities, $\textit{i.e.}$, audio and visual.
Therefore, when given such prompts $\hat{\mathbf{P}}^4$ and $\hat{\mathbf{P}}^5$, the uncertainty of the visual and audio prompt $\hat{\mathbf{P}}^6$ should also decrease.
Hence, we minimize the conditional entropy $H(\hat{\mathbf{P}}^6|\hat{\mathbf{P}}^4,\hat{\mathbf{P}}^5)$.
More generally, we aim to minimize the conditional entropy
$H(\hat{\mathbf{P}}^l|\hat{\mathbf{P}}^m,\hat{\mathbf{P}}^n)$, where $\hat{\mathbf{P}}^l,\hat{\mathbf{P}}^m,\hat{\mathbf{P}}^n \in \mathcal{P}_d$ and $l$, $m$, and $n$ are mutually distinct.
Similarly, we firstly define the joint probability distribution $P^{(l,m,n)} \in \mathbb{R}^{D\times D \times D}$ of $z_l$, $z_m$, and $z_n$ as follows:
\begin{equation}\nonumber
    P^{(l,m,n)}_{i,j,k}=\frac{1}{l}\sum_{t=1}^{l}\hat{\mathbf{P}}^l_{t,i}\hat{\mathbf{P}}^m_{t,j}\hat{\mathbf{P}}^n_{t,k}.
\end{equation}
The loss function between the $l$-th, $m$-th, and $n$-th prompts can be defined as:
\begin{equation}\nonumber
    \begin{aligned}  \mathcal{L}^{(l,m,n)}
    -\sum_{d_1=1}^{D}\sum_{d_2=1}^{D}\sum_{d_3=1}^{D}{}P^{(l,m,n)}_{d_1,d_2,d_3}\ln\frac{P^{(l,m,n)}_{d_1,d_2,d_3}}{P^{(l,m,n)}_{d_2,d_3}},
    \end{aligned}
\end{equation}
where $P^{(l,m,n)}_{d_2,d_3}$ is the marginal probability distribution by summing the first dimension of $P^{(l,m,n)}$.
The total loss function $\mathcal{L}_{cc}$ can be defined as follows:
\begin{equation}
    \mathcal{L}_{cc}=\mathcal{L}^{(3,4,5)}+\mathcal{L}^{(4,5,3)}+\mathcal{L}^{(5,3,4)}.
\end{equation}

\emph{Conditional Information Alignment}.
Given a multimodal combined prompt such as $\hat{\mathbf{P}}^6$ containing audio and visual information, for the prompt $\hat{\mathbf{P}}^4$ containing audio and text and the prompt $\hat{\mathbf{P}}^5$ containing visual and text, their shared text modality should remain consistent and aligned. 
Therefore, we maximize the conditional mutual information $I(\hat{\mathbf{P}}^4;\hat{\mathbf{P}}^5|\hat{\mathbf{P}}^6)$.
More generally, we aim to maximize $I(\hat{\mathbf{P}}_l;\hat{\mathbf{P}}^m|\hat{\mathbf{P}}^n)$, where $\hat{\mathbf{P}}^l,\hat{\mathbf{P}}^m,\hat{\mathbf{P}}^n \in \mathcal{P}_d$ and $l$, $m$, and $n$ are mutually distinct.
The loss function between $l$-th, $m$-th and $n$-th prompts can be defined as:
\begin{equation}\nonumber
\begin{aligned}
\hat{\mathcal{L}}^{(l,m,n)}
    =-\sum_{d_1=1}^{D}\sum_{d_2=1}^{D}\sum_{d_3=1}^{D}{}P^{(l,m,n)}_{d_1,d_2,d_3}\ln\frac{P^{(l,m,n)}_{d_1,d_2,d_3}P^{(l,m,n)}_{d_3}}{P^{(l,m,n)}_{d_1,d_3}P^{(l,m,n)}_{d_2,d_3}}.
\end{aligned}
\end{equation}
The total loss function $\mathcal{L}_{cia}$ can be defined as follows:
\begin{equation}
\mathcal{L}_{cia}=\hat{\mathcal{L}}^{(3,4,5)}+\hat{\mathcal{L}}^{(4,5,3)}+\hat{\mathcal{L}}^{(5,3,4)}.
\end{equation}

Finally, the total \emph{prompt contrastive learning} loss function $\mathcal{L}_{pcl}$ can be defined as:
\begin{equation}
    \mathcal{L}_{pcl}=\mathcal{L}_{mip}+\lambda_1\cdot \mathcal{L}_{cc}+\lambda_2\cdot\mathcal{L}_{cia},
\end{equation}
where $\lambda_1$ and $\lambda_2$ are trade-off parameters.

\begin{table*}[t]
\centering
\setlength{\tabcolsep}{5.2pt}
\begin{tabular}{c|c|cccccccccccc}
\toprule
\multirow{2}{*}{\textbf{Dataset}} & \multirow{2}{*}{\textbf{Method}}  & \multicolumn{2}{c}{\textbf{Task 1}} & \multicolumn{2}{c}{\textbf{Task 2}} & \multicolumn{2}{c}{\textbf{Task 3}}  & \multicolumn{2}{c}{\textbf{Task 4}} & \multicolumn{2}{c}{\textbf{Task 5}} & \multicolumn{2}{c}{\textbf{Task 6}}\\
\cmidrule(lr){3-4} \cmidrule(lr){5-6} \cmidrule(lr){7-8} \cmidrule(lr){9-10} \cmidrule(lr){11-12} \cmidrule(lr){13-14}
 & & \textbf{Acc-2} & \textbf{F1}& \textbf{Acc-2} & \textbf{F1} & \textbf{Acc-2} & \textbf{F1} & \textbf{Acc-2} & \textbf{F1} & \textbf{Acc-2} & \textbf{F1} & \textbf{Acc-2} & \textbf{F1} \\
\midrule
\multirow{9}{*}{\rotatebox[origin=c]{270}{CMU-MOSI}} & Self-MM & 65.70 & 64.81 & 67.22 & 66.63 & 65.55 & 65.00 & 68.70 & 67.33 & 66.82 & 66.23 & 64.43 & 63.17 \\
 & CubeMLP & \underline{69.41} & \underline{69.44} & \underline{69.97} & 69.88 & \underline{69.61} & \underline{69.71} & 68.50 & 68.61 & \underline{70.17} & \underline{70.27} & 66.77 & 66.42 \\
 & DMD & 67.68 & 67.06 & 69.92 & \underline{69.91} & 69.46 & 69.50 & 67.53 & 67.05 & 69.87 & 69.96 & 67.83 & \underline{67.94} \\
 & DLF & 67.27 & 66.81 & 69.51 & 69.38 & 68.80 & 67.97 & \underline{69.77} & \underline{69.89} & 70.02 & 69.66 & 65.85 & 65.58 \\
 & TFR-Net & 60.06 & 57.77 & 47.51 & 42.94 & 50.31 & 42.17 & 55.44 & 49.83 & 53.86 & 50.73 & 49.95 & 44.55 \\
 & MPLMM & 55.03 & 48.38 & 53.71 & 46.35 & 59.20 & 55.02 & 58.64 & 55.80 & 56.40 & 50.50 & 65.55 & 64.53 \\
 & MFMB-Net & 67.04 & 66.01 & 68.57 & 68.31 & 69.14 & 68.72 & 69.28 & 68.12 & 68.03 & 67.43 & \underline{67.96} & 67.79 \\
 & LNLN & 66.06 & 65.17 & 66.57 & 65.64 & 67.79 & 67.32 & 68.45 & 68.14 & 67.63 & 67.04 & 67.02 & 66.46 \\
 & CMPL & \textbf{75.36} & \textbf{75.43} & \textbf{75.56} & \textbf{75.55} & \textbf{75.15} & \textbf{75.21} & \textbf{76.88} & \textbf{76.80} & \textbf{76.67} & \textbf{76.79} & \textbf{73.63} & \textbf{73.63} \\
\midrule
\multirow{9}{*}{\rotatebox[origin=c]{270}{SIMS-V2}} & Self-MM & 66.28 & 64.83 & 66.34 & 64.11 & 63.73 & 62.65 & 66.92 & 64.41 & 64.22 & 62.77 & 64.18 & 62.89 \\
 & CubeMLP & 67.57 & 67.13 & 66.02 & 65.60 & 65.25 & 60.05 & 65.12 & 59.95 & 64.57 & 59.40 & 63.06 & 57.91 \\
 & DMD & 67.47 & 66.52 & 67.60 & 66.94 & 68.60 & 68.39 & 67.89 & 66.77 & 68.67 & 68.25 & \underline{71.24} & \underline{70.73} \\
 & DLF & 68.31 & 67.16 & 70.31 & 69.83 & 71.02 & 70.79 & 70.86 & 70.48 & 68.47 & 67.43 & 70.18 & 69.84 \\
 & TFR-Net & 65.96 & 65.42 & 65.05 & 65.18 & 61.77 & 61.61 & 64.80 & 64.24 & 64.22 & 60.44 & 60.80 & 55.45 \\
 & MPLMM & 62.67 & 61.29 & 68.12 & 67.89 & 62.99 & 59.64 & 64.02 & 63.34 & 65.89 & 63.80 & 61.77 & 60.75 \\
 & MFMB-Net & 70.74 & \underline{70.75} & 67.96 & 67.67 & 71.12 & \underline{70.98} & 69.70 & 69.08 & 66.96 & 66.85 & 70.80 & 70.46 \\
 & LNLN & \underline{71.02} & 70.30 & \underline{70.63} & \underline{70.44} & \underline{71.50} & 70.59 & \underline{71.47} & \underline{70.90} & \underline{71.47} & \underline{70.94} & 70.86 & 70.40 \\
 & CMPL & \textbf{77.95} & \textbf{78.00} & \textbf{76.92} & \textbf{77.02} & \textbf{78.34} & \textbf{78.36} & \textbf{77.69} & \textbf{77.75} & \textbf{78.21} & \textbf{78.20} & \textbf{76.72} & \textbf{76.70} \\
\midrule
\multirow{9}{*}{\rotatebox[origin=c]{270}{CMU-MOSEI}} & Self-MM & 74.50 & 72.69 & 74.53 & 73.02 & 74.02 & 72.21 & 74.42 & 72.60 & 75.06 & 72.79 & 73.78 & 70.83 \\
 & CubeMLP & 74.63 & 73.02 & 71.01 & 64.43 & 70.88 & 65.10 & 74.38 & 72.86 & 66.39 & 61.60 & 67.19 & 62.74 \\
 & DMD & \underline{75.73} & \underline{74.36} & 75.35 & \underline{74.47} & 74.51 & 73.70 & \underline{75.60} & \underline{73.77} & 75.08 & 73.38 & 74.24 & 73.59 \\
 & DLF & 75.72 & 74.14 & \underline{75.50} & 74.22 & \underline{75.42} & \underline{73.90} & 74.98 & 73.64 & 75.30 & \underline{73.95} & \underline{74.99} & \underline{73.96} \\
 & TFR-Net & 71.44 & 68.50 & 73.43 & 71.55 & 71.71 & 66.74 & 73.09 & 71.15 & 70.06 & 67.15 & 68.35 & 60.86 \\
 & MPLMM & 72.17 & 71.23 & 70.78 & 70.11 & 70.06 & 69.36 & 73.17 & 71.76 & 69.99 & 69.99 & 71.42 & 70.39 \\
 & MFMB-Net & 74.09 & 70.57 & 72.94 & 68.79 & 74.20 & 71.94 & 73.78 & 70.34 & 74.11 & 71.25 & 72.36 & 68.76 \\
 & LNLN & 75.43 & 74.19 & 75.20 & 73.98 & 75.19 & 73.51 & 75.22 & 73.27 & \underline{75.35} & 73.46 & 74.72 & 73.14 \\
 & CMPL & \textbf{77.39} & \textbf{76.59} & \textbf{77.41} & \textbf{76.88} & \textbf{77.27} & \textbf{76.79} & \textbf{76.28} & \textbf{75.86} & \textbf{76.94} & \textbf{76.29} & \textbf{75.94} & \textbf{74.92} \\
\bottomrule
\end{tabular}
\caption{The performance of different methods on various datasets under six tasks. The best and second-best results are marked in bold and underlined, respectively.}
\label{tab:multi_dataset_results}
\end{table*}

\section{Experiments}

\subsection{Experiment Setting}
\subsubsection{Datasets.}
We conducted experiments on three widely used datasets, including CMU-MOSI~\cite{zadeh2016multimodal}, CMU-MOSEI~\cite{zadeh2018multimodal}, and SIMS-V2~\cite{yu2020ch}.  
The CMU-MOSI dataset contains a total of 2,199 video clips, each manually annotated with sentiment scores ranging from strongly negative to strongly positive (-3 to 3). 
The CMU-MOSEI dataset consists of 22,856 video clips, covering a broader range of topics compared to CMU-MOSI, with sentiment labels also annotated on the same scale (-3 to 3).
SIMS-V2 is a Chinese multimodal sentiment analysis dataset containing 4,403 video clips, where sentiment values are labeled from -1 to 1.

\begin{table}[t]
\centering
\setlength{\tabcolsep}{17pt}
\begin{tabular}{ccc}
\toprule
\textbf{Task No.} & \textbf{Training} & \textbf{Test} \\
\midrule
Task 1 & $\{S_1,S_2,S_3,S_4,S_5\}$ & $\mathcal{S}$\\
Task 2 & $\{S_1,S_2,S_3,S_4,S_6\}$ & $\mathcal{S}$\\
Task 3 & $\{S_1,S_2,S_3,S_5,S_6\}$ & $\mathcal{S}$\\
Task 4 & $\{S_1,S_2,S_3,S_4\}$ & $\mathcal{S}$\\
Task 5 & $\{S_1,S_2,S_3,S_5\}$ & $\mathcal{S}$\\
Task 6 & $\{S_1,S_2,S_3,S_6\}$ & $\mathcal{S}$\\
Task 7 & $\mathcal{S}$ & $\mathcal{S}$\\
\bottomrule
\end{tabular}
\caption {The cases of modality combinations in the training and test sets across the seven tasks.}
\label{tab:task}
\vspace*{-0.1in}
\end{table}

\begin{figure*}[t]
    \centering
    \includegraphics[width=0.95\linewidth]{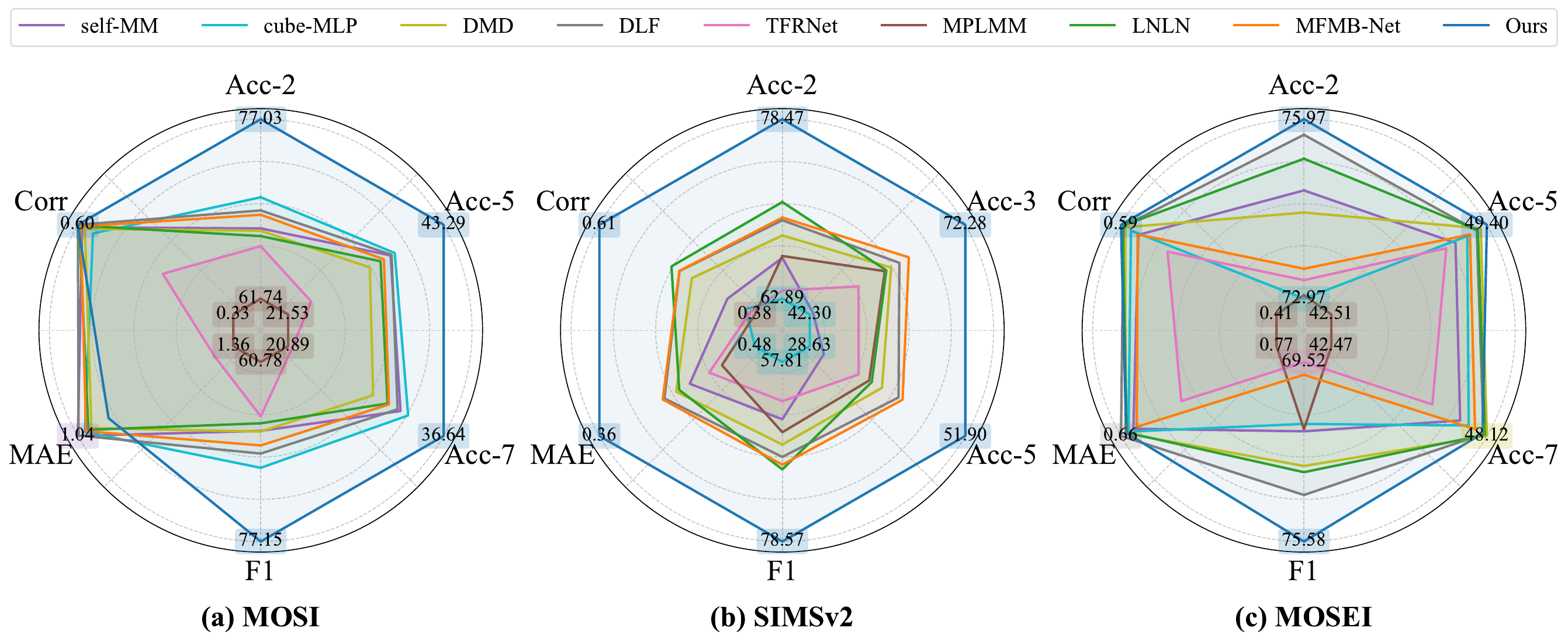}
    \caption{The performance of seven methods across six evaluation metrics on three datasets under Task 7. The center of the radar chart represents the worst results, and the outermost vertices correspond to the best results.}
    \label{fig:radar}
   \vspace*{-0.1in}
\end{figure*}

\subsubsection{Metrics.}
Due to differences in labels across datasets, we employ different evaluation metrics for different datasets.  
For CMU-MOSI and CMU-MOSEI, we adopt binary accuracy (Acc-2), five-category accuracy (Acc-5), seven-category accuracy (Acc-7), F1 score, mean absolute error (MAE), and Pearson correlation (Corr) as evaluation metrics.  
For SIMS-V2, we use  Acc-2, three-category accuracy (Acc-3), Acc-5, F1, MAE, and Corr.
Here, Acc-2 and F1 follow the non-positive/positive standard.

\subsubsection{Baselines.}
In our experiments, we compare with six state-of-the-art methods, including modality-complete methods: Self-MM, Cube-MLP, DMD, and DLF; and modality-missing methods: TFRNet, MPLMM, MFMB-Net and LNLN.
For methods that require complete modalities during training, we fill in missing modalities with zero.
For all methods, we keep the parameters recommended in their original papers or released codes.

\subsubsection{Implementation Details.}
We conduct experiments on the Ubuntu 20.04 system with an Intel(R) Xeon(R) Gold 6326 CPU @ 2.90GHz and a single NVIDIA A40. 
We adopt Qwen1.5-1.8b~\cite{bai2023qwen} as the backbone.
For the training process, we use the Adam optimizer with a learning rate of $1\times10^{-4}$. 
For reliability, we perform three independent runs for each experiment and report the average results.
More details can be found in Appendix \ref{sec:imple}.


\subsection{Main Results}
In the experiment, we design seven experimental scenarios, \textit{i.e.}, seven different tasks, as shown in Table~\ref{tab:task}, where $S_i$ represents the $i$-th modality combination and $\mathcal{S}=\{S_1,S_2,\cdots,S_7\}$. 
For all scenarios, the testing phase has all modality combinations (\textit{i.e.}, $\mathcal{S}$).

Table~\ref{tab:multi_dataset_results} presents the results of seven methods across three datasets under seven different scenarios. From Table~\ref{tab:multi_dataset_results}, it can be observed that our method achieves the best performance in almost all scenarios.
Compared to the second-best method, LNLN, our approach improves accuracy by an average of 5.57\%.
Notably, on the CMU-MOSI dataset, our method outperforms LNLN by an average of 13.26\% in F1-score.
In particular, under the Task 6 scenario, the F1-score improvement reaches 16.24\%.
This is because our method, \textsf{CMPL}, leverages label-guided contrastive feature learning to effectively capture multimodal consistency and the semantic structure of instances, employing three prompt contrastive learning strategies to enhance the model's ability to learn from each modality combination and generalize to unseen modality combinations.

Furthermore, to further validate the effectiveness of our method, we conduct experiments under the scenario where both the training and test sets contain all seven modality combinations (\textit{i.e.}, ${S}_7$), as shown in Figure~\ref{fig:radar}. 
The results demonstrate that our approach still outperforms others in most cases, highlighting its superiority and robustness.
More results can be found in Appendix \ref{sec:add_results}.


\begin{table}[t]
\setlength{\tabcolsep}{1.00mm}
\centering
\begin{tabular}{c|cccccc}
\toprule
\textbf{Models} & \textbf{Acc-2} & \textbf{F1} & \textbf{MAE} & \textbf{Corr} & $T_{train}$ & $T_{test}$ \\
\midrule
Qwen1.5-1.8B & 75.36 & 75.43 & \underline{1.068} & 0.579 & \textbf{6.1m} & \textbf{6.0s} \\
Llama3.2-3B  & 75.71 & 75.51 & 1.096 & \underline{0.581} & \underline{9.0m} & \underline{8.4s} \\
Llama-2-7B   & \underline{76.60} & \underline{76.53} & 1.087 & 0.574 & 25.8m & 15.1s \\
Qwen3-8B     & \textbf{77.90} & \textbf{78.03} & \textbf{0.989} & \textbf{0.636} & 18.6m & 16.8s \\
\bottomrule
\end{tabular}
\caption{The performance with different LLMs on the CMU-MOSI dataset under Task 4, where the units for training time and testing time are minutes (m) and seconds (s), respectively.}
\label{tab:df_llm}
\vspace*{-0.1in}
\end{table}

\subsection{Comparable Results with Different LLMs}
To further validate the effectively of \textsf{CMPL}, we conduct experiments using different LLMs of varying scales (\textit{i.e.}, Qwen1.5-1.8B, Llama3.2-3B~\cite{dubey2024llama}, Llama-2-7B~\cite{touvron2023llama}, and Qwen3-8B~\cite{yang2025qwen3}) as the backbone under the Task 1 on the CMU-MOSI dataset. 
The performance and time of training and testing are presented in Table~\ref{tab:df_llm}.

As shown, models with larger parameter sizes generally achieve higher performance than smaller ones.
For instance, Qwen3-8B achieves a 9.84\% improvement in Corr compared to Qwen1.5-1.8B, which can be attributed to its greater learning capacity and ability to capture more nuanced knowledge. 
However, for computational resources, 
Qwen3-8B requires over three times the training time of Qwen1.5-1.8B. 
Given that Qwen1.5-1.8B offers a favorable balance between resource efficiency and performance, it serves as a practical and cost-effective choice for common deployment.

\subsection{Ablation Study}
To validate the effectiveness of each module in our method, we conduct experiments on the CMU-MOSI and SIMS-V2 datasets under the Task 4 scenario. 
We systematically remove each module and observe the model's performance changes.
The ablation results are presented in Table~\ref{tab:ablation_results}.

The ablation results demonstrate that removing any module leads to performance degradation, while the model achieves its optimal performance when all modules are intact. 
Specifically, on CMU-MOSI dataset, removing the LCFL module results in a 2.79\% decrease in Corr, while removing the PCL module causes a 2.94\% increase in MAE on SIMS-V2 dataset.
What's more, the removal of the MPL module causes both a 2.64\% drop in Acc-2 and a substantial 2.83\% in F1 on CMU-MOSI dataset.
These results confirm that each module plays a critical role, validating the contributions of each module to the model's effectiveness.
More additional ablation results can be found in Appendix \ref{sec:add_results}.

\begin{table}[t]
\centering
\setlength{\tabcolsep}{4pt}
\begin{tabular}{c|c|cccc}
\toprule
\textbf{Datasets} & \textbf{Methods} & \textbf{Acc-2} & \textbf{F1} & \textbf{MAE} & \textbf{Corr} \\
\midrule
\multirow{4}{*}{CMU-MOSI} 
 & w/o LCFL & 75.25 & 75.02 & 1.107 & 0.574 \\
 & w/o MPL & 74.90 & 74.69 & \underline{1.089} & 0.574 \\
 & w/o PCL & \underline{75.96} & \underline{75.85} & 1.100 & \underline{0.586} \\
 & CMPL & \textbf{76.88} & \textbf{76.80} & \textbf{1.079} & \textbf{0.590} \\
\midrule
\multirow{4}{*}{SIMS-V2} 
 & w/o LCFL & 76.27 & 76.31 & 0.375 & 0.569 \\
 & w/o MPL & \underline{77.24} & \underline{77.11} & \underline{0.365} & \underline{0.574} \\
 & w/o PCL & 76.47 & 76.49 & 0.374 & \underline{0.574} \\
 & CMPL & \textbf{77.69} & \textbf{77.75} & \textbf{0.363} & \textbf{0.593} \\
\bottomrule
\end{tabular}
\caption{The ablation study on both CMU-MOSI and SIMS-V2 datasets under the Task 4.}
\label{tab:ablation_results}
\vspace*{-0.1in}
\end{table}

\section{Conclusion}
In this paper, we propose a novel model named \textsf{CMPL} to address the task of incomplete multimodal sentiment analysis with the unseen modality combination. 
We introduce a \emph{label-guided contrastive feature learning} mechanism to maintain multimodal consistency while preserving the structural relationships among data. 
Furthermore, we develop a \emph{mixed prompts learning} mechanism incorporating the \emph{prompt contrastive learning} strategies, which effectively enhances the model's comprehension of diverse modal combinations and improves its generalization capability to unseen modal combinations. 
Extensive experiments validate the effectiveness of our approach.

\bibliography{aaai2027}


\appendix
\clearpage

\setcounter{secnumdepth}{1}

\begin{table}[t]
  \centering
  \setlength{\tabcolsep}{9.1pt}
  \begin{tabular}{ccc}
    \toprule
    \textbf{No.} & \textbf{\{Available, Missing\}} & \textbf{Available Data} \\
    \midrule
    $S_1$ & \{(t),\ (a, v)\} & $\mathbf{x}=\{\mathbf{x}^t\}$\\ 
    $S_2$ & \{(a),\ (t, v)\} & $\mathbf{x}=\{\mathbf{x}^a\}$\\ 
    $S_3$ & \{(v),\ (t, a)\} & $\mathbf{x}=\{\mathbf{x}^v\}$\\ 
    $S_4$ & \{(t,\ a),\ (v)\} & $\mathbf{x}=\{\mathbf{x}^t, \mathbf{x}^a\}$\\ 
    $S_5$ & \{(t,\ v),\ (a)\} & $\mathbf{x}=\{\mathbf{x}^t, \mathbf{x}^v\}$\\ 
    $S_6$ & \{(a,\ v),\ (t)\} & $\mathbf{x}=\{\mathbf{x}^a, \mathbf{x}^v\}$\\ 
    $S_7$ & \{(t,\ a,\ v),\ ()\} & $\mathbf{x}=\{\mathbf{x}^t, \mathbf{x}^a, \mathbf{x}^v\}$\\ 
    \bottomrule
  \end{tabular}
    \caption{The seven modality combination cases.}
  \label{tab:combination}
\end{table}

\begin{table}[t]
\centering
\setlength{\tabcolsep}{3.pt}
\begin{tabular}{cccc}
\toprule
\textbf{Parameters} & \textbf{CMU-MOSI} & \textbf{SIMS-V2} & \textbf{CMU-MOSEI}\\
\midrule
Learning Rate & 1e-4 & 1e-4 & 1e-4\\
Batch Size & 64 & 64 & 64\\
$\alpha$, $\beta$ & 1, 1e-2 & 1, 1e-1 & 1, 1e-2\\
$\lambda_1$, $\lambda_2$ & 10, 1 & 1e-1, 1 & 10, 1\\
Pool Size & 32 & 32 & 32\\
Optimizer & Adam & Adam & Adam\\
\bottomrule
\end{tabular}
\caption {Experimental parameters of three datasets.}
\label{tab:Implementation}
\end{table}

\begin{table}[t]
\centering
\setlength{\tabcolsep}{9pt}
\begin{tabularx}{\linewidth}{cX}
\toprule
\textbf{Dataset} & \multicolumn{1}{c}{\textbf{Initial Text of Prompt}} \\  
\midrule
\multirow{3}{*}{CMU-MOSI} & Please predict the sentiment intensity of the \textless content\textgreater with in the range [-3.0, +3.0]. \\
\addlinespace 
\multirow{3}{*}{SIMS-V2} & Please predict the sentiment intensity of the \textless content\textgreater with in the range [-1.0, +1.0]. \\
\addlinespace 
\multirow{3}{*}{CMU-MOSEI} & Please predict the sentiment intensity of the \textless content\textgreater with in the range [-3.0, +3.0]. \\
\bottomrule
\end{tabularx}
\caption {The initial text of the modality combination prompts for three datasets.}
\label{tab:prompt}
\end{table}

\section{Task Description}\label{sec:task_desc}
In the IMSAUMC task, given three modalities, excluding the case where all modalities are absent, there are a total of $2^3 - 1 = 7$ modality combinations. 
The corresponding available and missing states of the modalities, as well as the available data, are shown in Table~\ref{tab:combination}.

\section{Implementation Details}\label{sec:imple}
Some parameters involved in the experiment are shown in Table~\ref{tab:Implementation}.
Table~\ref{tab:prompt} displays the initial text of prompts in three datasets, where ``$\textless $content$\textgreater$'' will be replaced according to different modality combinations. 
For instance, when the input is an audio-vision modality combination, ``$\textless $content$\textgreater$'' will be replaced with ``audio and vision content'', and similarly for other cases.


\begin{table}[t]
\centering
\setlength{\tabcolsep}{4pt}
\begin{tabular}{c|c|cccc}
\toprule
\textbf{Datasets} & \textbf{Methods} & \textbf{Acc-2} & \textbf{F1} & \textbf{MAE} & \textbf{Corr} \\
\midrule
\multirow{4}{*}{CMU-MOSI} 
 & w/o MIP & 76.02 & 72.47 & 1.181 & 0.549 \\
 & w/o CC & \underline{76.68} & \underline{76.53} & \underline{1.088} & \underline{0.577} \\
 & w/o CIA & 76.32 & 76.42 & 1.108 & 0.571 \\
 & CMPL & \textbf{76.88} & \textbf{76.80} & \textbf{1.079} & \textbf{0.590} \\
\midrule
\multirow{4}{*}{SIMS-V2} 
 & w/o MIP & 76.89 & 76.77 & 0.370 & 0.580 \\
 & w/o CC & \underline{77.53} & \underline{77.52} & 0.365 & 0.587 \\
 & w/o CIA & 76.72 & 76.83 & \underline{0.364} & \underline{0.590} \\
 & CMPL & \textbf{77.69} & \textbf{77.75} & \textbf{0.363} & \textbf{0.593} \\
\bottomrule
\end{tabular}
\caption{Further ablation study on both CMU-MOSI and SIMS-V2 datasets under the Task 4.}
\label{tab:further_ablation_results}
\end{table}


\section{Additional Experimental Results}\label{sec:add_results}
\subsection{Parameter Analysis}
\begin{figure*}[t]
    \centering

    \begin{subfigure}{0.4\linewidth}
        \centering
        \includegraphics[width=\linewidth]{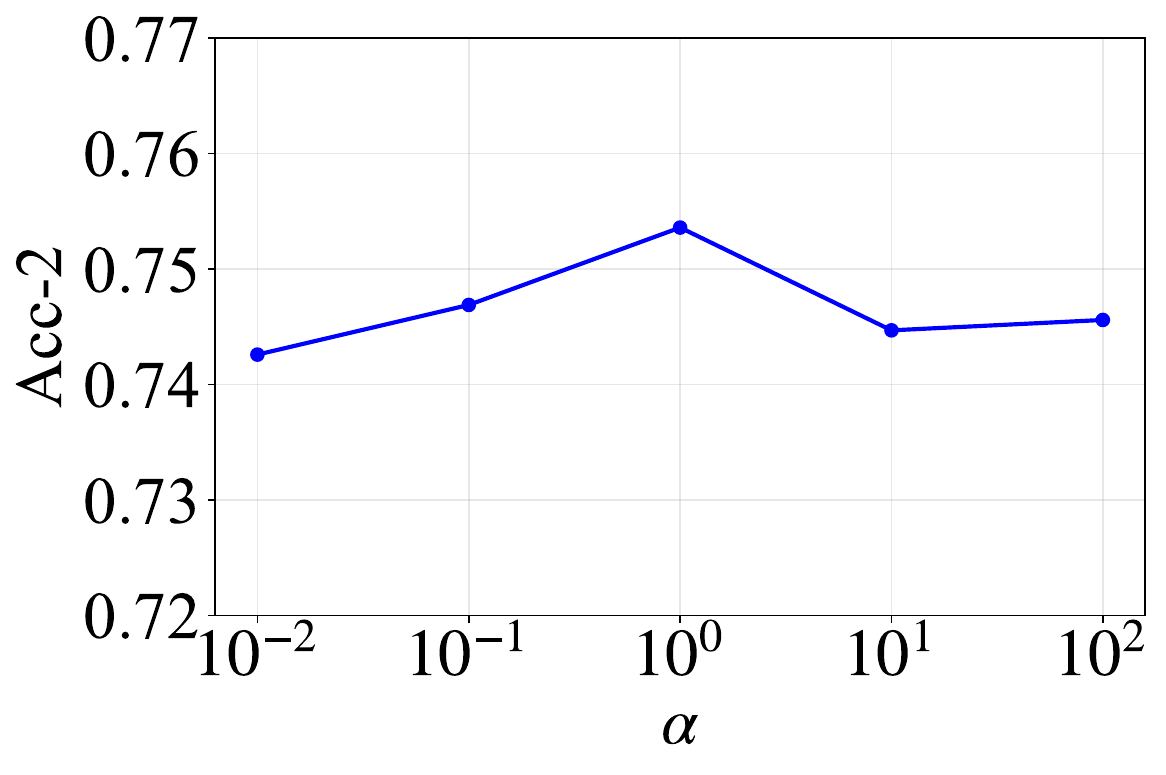}
        \caption{Parameter Analysis of $\alpha$}
    \end{subfigure}
    \hspace{20pt}
    \begin{subfigure}{0.4\linewidth}
        \centering
        \includegraphics[width=\linewidth]{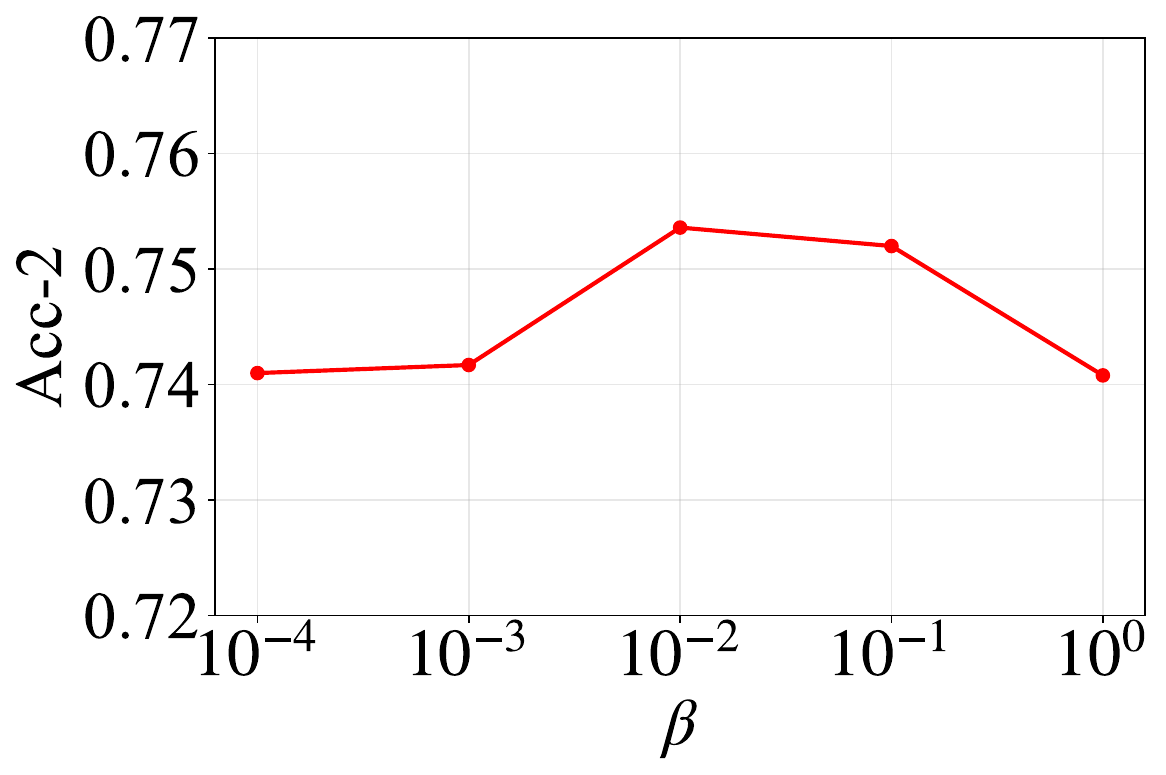}
        \caption{Parameter Analysis of $\beta$}
    \end{subfigure}

    \begin{subfigure}{0.4\linewidth}
        \centering
        \includegraphics[width=\linewidth]{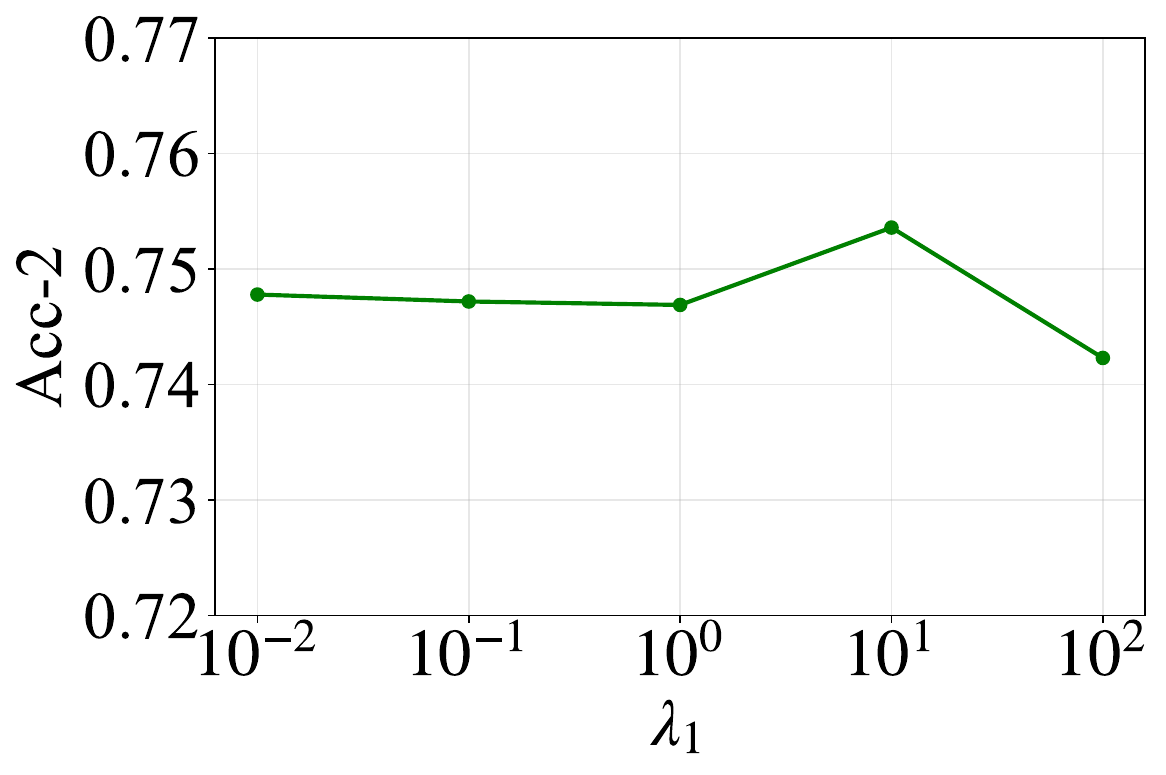}
        \caption{Parameter Analysis of $\lambda_1$}
    \end{subfigure}
    \hspace{20pt}
    \begin{subfigure}{0.4\linewidth}
        \centering
        \includegraphics[width=\linewidth]{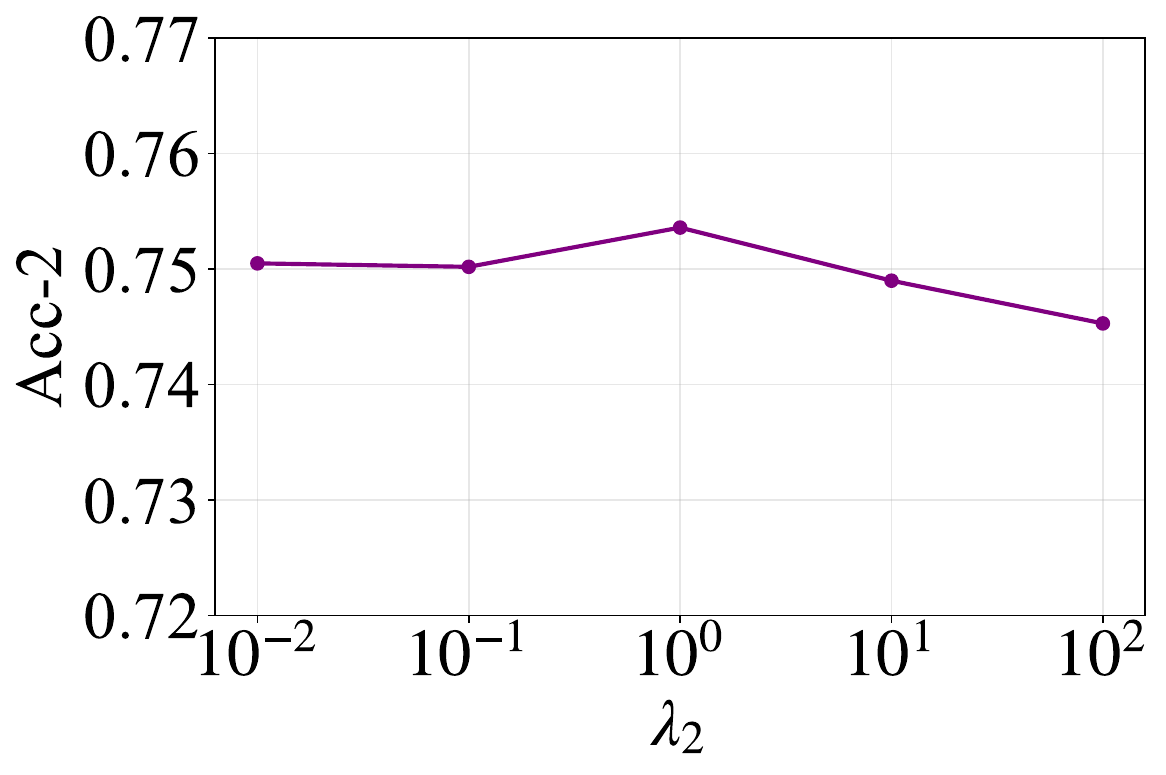}
        \caption{Parameter Analysis of $\lambda_2$}
    \end{subfigure}

    \caption{Parameter analysis on the CMU-MOSI dataset.}
    \label{fig:Parameter}
\end{figure*}

We conduct a parameter analysis on the CMU-MOSI dataset for the key hyperparameters used in CMPL. 
Specifically, we examine the parameter $\alpha$, which controls the LCFL loss when the PCL strategy is disabled. 
We then analyze the parameters $\beta$, $\lambda_1$, and $\lambda_2$, which regulate the MIP, CC, and CIA components in the PCL strategy.
The parameter analysis results are presented in Figure~\ref{fig:Parameter}.

The results show that all four parameters lead to certain performance variations when ranging from $1e-2$ to $1e2$. 
However, the overall performance of the model is not highly sensitive to these changes. 
The model achieves its best performance when $\alpha=1$, $\beta=1e-2$, $\lambda_1=10$, and $\lambda_2=1$.

\subsection{Further Ablation Study}
To further validate the contribution of each strategy within PCL, we conducted additional ablation studies by individually removing the MIP, CC, and CIA strategies. 
While the ablation results have already been presented in the main text in graphical form, we provide the detailed numerical results in Table~\ref{tab:further_ablation_results}. 
As shown, removing any single strategy leads to a performance drop, which further demonstrates the effectiveness of each component in our PCL framework.

\begin{figure*}[t]
    \centering
    \includegraphics[width=0.7\linewidth]{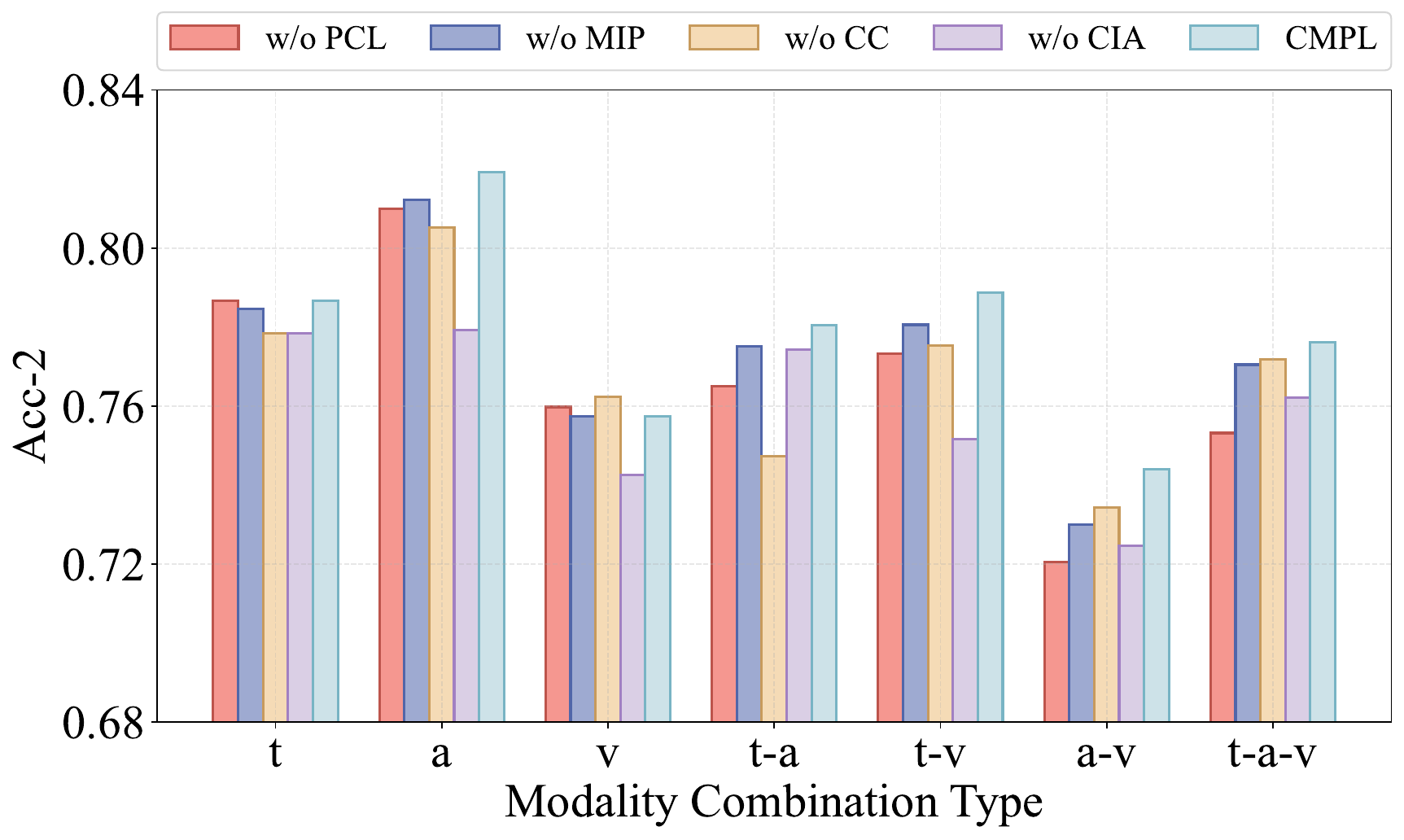}
    \caption{The model performance for various modality combinations on SIMS-V2 dataset under Task 4.}
    \label{fig:acc_ablation}
\end{figure*}

To further validate the effectiveness of PCL on the IMSAUMC task, we additionally present its performance for each modality combination in the test set under Task 4 on the SIMS-V2 dataset.
The results are shown in Figure~\ref{fig:acc_ablation}, where ``t'', ``a'', and ``v'' denote text, audio, and visual modality, while ``t-a'' indicates the text-audio modality combination and similarly for others.
Here, ``w/o MIP'', ``w/o CC'', ``w/o CIA'' indicate the absence of each strategy in PCL.
As observed, in the absence of any prompt learning strategies, the model performs poorly on unseen modality combinations, with an average accuracy drop of approximately 3.12\% compared to CMPL.
The incorporation of the PCL module demonstrates additional performance gains in audio-vision (a-v), and text-audio-vision (t-a-v) modality combinations.
Furthermore, when the MIP, CC, and CIA strategies are successively removed from the CMPL model, the performance decreases by about 1.32\%, 0.94\%, and 2.26\% on unseen modality combinations, respectively, which indicates that each strategy plays a critical role in improving the model's generalization.
The model achieves optimal performance when all three strategies are incorporated. These results demonstrate the effectiveness and necessity of each strategy in PCL.



\subsection{Detailed Experimental Results}
Table~\ref{tab:multi_dataset_results_1} to Table~\ref{tab:multi_dataset_results_7} present detailed experimental results of the seven methods across seven tasks on three datasets, evaluated by six metrics. 
As shown in the results, our model consistently achieves superior performance over baseline methods in most scenarios, clearly validating the effectiveness and superiority of our approach.

\begin{table*}[t]
\centering
\setlength{\tabcolsep}{6pt}
\begin{tabular}{c|c|ccccccccc}
\toprule
\textbf{Dataset} & \textbf{Metrics} & \textbf{self-MM} & \textbf{cube-MLP} & \textbf{DMD} & \textbf{DLF} & \textbf{TFRNet} & \textbf{MPLMM} & \textbf{MFMB-Net} & \textbf{LNLN} & \textbf{Ours}\\
\midrule
\multirow{6}{*}{\rotatebox[origin=c]{270}{CMU-MOSI}} & Acc-2 & 65.70 & \underline{69.41} & 67.68 & 67.27 & 60.06 & 55.03 & 67.04 & 66.06 & \textbf{75.36} \\
 & Acc-5 & 33.82 & \underline{35.03} & 33.63 & 31.58 & 21.14 & 18.71 & 32.19 & 29.30 & \textbf{40.82} \\
 & Acc-7 & 29.74 & \underline{30.95} & 29.88 & 28.96 & 20.07 & 18.61 & 27.68 & 27.11 & \textbf{34.50} \\
 & F1 & 64.81 & \underline{69.44} & 67.06 & 66.81 & 57.77 & 48.38 & 66.01 & 65.17 & \textbf{75.43} \\
 & MAE & 1.088 & \underline{1.078} & 1.086 & 1.078 & 1.297 & 1.378 & 1.127 & 1.118 & \textbf{1.068} \\
 & Corr & 0.572 & 0.569 & \textbf{0.587} & 0.579 & 0.378 & 0.294 & 0.568 & \underline{0.586} & 0.579 \\
\midrule
\multirow{6}{*}{\rotatebox[origin=c]{270}{SIMS-V2}} & Acc-2 & 66.28 & 67.57 & 67.47 & 68.31 & 65.96 & 62.67 & 70.74 & \underline{71.02} & \textbf{77.95} \\
 & Acc-3 & 48.32 & \underline{59.83} & 54.06 & 55.67 & 52.74 & 55.80 & 59.27 & 59.06 & \textbf{71.44} \\
 & Acc-5 & 33.11 & \underline{40.78} & 36.94 & 38.49 & 35.82 & 36.69 & 40.44 & 40.72 & \textbf{50.45} \\
 & F1 & 64.83 & 67.13 & 66.52 & 67.16 & 65.42 & 61.29 & \underline{70.75} & 70.30 & \textbf{78.00} \\
 & MAE & 0.437 & 0.433 & 0.420 & 0.420 & 0.427 & 0.486 & \underline{0.410} & 0.415 & \textbf{0.358} \\
 & Corr & 0.429 & 0.449 & 0.461 & 0.478 & 0.444 & 0.288 & 0.499 & \underline{0.506} & \textbf{0.611} \\
\midrule
\multirow{6}{*}{\rotatebox[origin=c]{270}{CMU-MOSEI}} & Acc-2 & 74.50 & 74.63 & \underline{75.73} & 75.72 & 71.44 & 72.17 & 74.09 & 75.43 & \textbf{77.39} \\
 & Acc-5 & 48.94 & 46.96 & 49.11 & 48.33 & 44.86 & 42.28 & \underline{49.59} & 48.55 & \textbf{49.68} \\
 & Acc-7 & 47.84 & 45.90 & 48.22 & 47.33 & 43.54 & 42.13 & \textbf{48.48} & 47.63 & \underline{48.30} \\
 & F1 & 72.69 & 73.02 & \underline{74.36} & 74.14 & 68.50 & 71.23 & 70.57 & 74.19 & \textbf{76.59} \\
 & MAE & 0.664 & 0.696 & \textbf{0.659} & 0.663 & 0.741 & 0.773 & 0.664 & 0.662 & \underline{0.660} \\
 & Corr & 0.578 & 0.551 & 0.593 & 0.591 & 0.544 & 0.407 & 0.581 & \textbf{0.595} & \underline{0.593} \\
\bottomrule
\end{tabular}
\caption {The performance of different methods on three datasets under Task 1. The best and second-best results are marked in bold and underlined, respectively.}
\label{tab:multi_dataset_results_1}
\end{table*}

\begin{table*}[t]
\centering
\setlength{\tabcolsep}{6pt}
\begin{tabular}{c|c|ccccccccc}
\toprule
\textbf{Dataset} & \textbf{Metrics} & \textbf{self-MM} & \textbf{cube-MLP} & \textbf{DMD} & \textbf{DLF} & \textbf{TFRNet} & \textbf{MPLMM} & \textbf{MFMB-Net} & \textbf{LNLN} & \textbf{Ours}\\
\midrule
\multirow{6}{*}{\rotatebox[origin=c]{270}{CMU-MOSI}} & Acc-2 & 67.22 & \underline{69.97} & 69.92 & 69.51 & 47.51 & 53.71 & 68.57 & 66.57 & \textbf{75.56} \\
 & Acc-5 & \underline{34.11} & 32.80 & 33.72 & 34.01 & 17.83 & 21.52 & 33.73 & 27.01 & \textbf{42.32} \\
 & Acc-7 & \underline{30.32} & 29.49 & 28.72 & 29.59 & 17.78 & 20.89 & 29.32 & 25.22 & \textbf{35.23} \\
 & F1 & 66.63 & 69.88 & \underline{69.91} & 69.38 & 42.94 & 46.35 & 68.31 & 65.64 & \textbf{75.55} \\
 & MAE & \textbf{1.062} & 1.078 & 1.075 & 1.085 & 1.480 & 1.340 & \underline{1.070} & 1.194 & 1.121 \\
 & Corr & 0.574 & 0.583 & 0.589 & \underline{0.593} & 0.139 & 0.326 & \textbf{0.596} & 0.551 & 0.586 \\
\midrule
\multirow{6}{*}{\rotatebox[origin=c]{270}{SIMS-V2}} & Acc-2 & 66.34 & 66.02 & 67.60 & 70.31 & 65.05 & 68.12 & 67.96 & \underline{70.63} & \textbf{76.92} \\
 & Acc-3 & 42.65 & 54.64 & 50.23 & 58.80 & 45.26 & \underline{61.38} & 57.62 & 58.41 & \textbf{72.11} \\
 & Acc-5 & 30.11 & 37.11 & 34.78 & 39.88 & 30.53 & \underline{40.14} & 39.57 & 39.33 & \textbf{51.84} \\
 & F1 & 64.11 & 65.60 & 66.94 & 69.83 & 65.18 & 67.89 & 67.67 & \underline{70.44} & \textbf{77.02} \\
 & MAE & 0.435 & 0.447 & 0.419 & 0.419 & 0.450 & 0.446 & 0.415 & \underline{0.414} & \textbf{0.376} \\
 & Corr & 0.421 & 0.404 & 0.459 & 0.511 & 0.399 & 0.419 & 0.495 & \underline{0.519} & \textbf{0.592} \\
\midrule
\multirow{6}{*}{\rotatebox[origin=c]{270}{CMU-MOSEI}} & Acc-2 & 74.53 & 71.01 & 75.35 & \underline{75.50} & 73.43 & 70.78 & 72.94 & 75.20 & \textbf{77.41} \\
 & Acc-5 & 48.84 & 44.13 & \underline{49.05} & 48.20 & 46.92 & 41.59 & 47.62 & \textbf{49.06} & 48.82 \\
 & Acc-7 & 47.82 & 43.62 & \textbf{48.16} & 47.33 & 46.07 & 41.59 & 46.93 & \underline{48.02} & 47.66 \\
 & F1 & 73.02 & 64.43 & \underline{74.47} & 74.22 & 71.55 & 70.11 & 68.79 & 73.98 & \textbf{76.88} \\
 & MAE & 0.667 & 0.751 & \textbf{0.663} & \underline{0.664} & 0.694 & 0.790 & 0.681 & 0.664 & 0.667 \\
 & Corr & 0.579 & 0.536 & 0.586 & \underline{0.591} & 0.551 & 0.394 & 0.554 & \textbf{0.595} & 0.587 \\
\bottomrule
\end{tabular}
\caption {The performance of different methods on three datasets under Task 2. The best and second-best results are marked in bold and underlined, respectively.}
\label{tab:multi_dataset_results_2}
\end{table*}

\begin{table*}[t]
\centering
\setlength{\tabcolsep}{6pt}
\begin{tabular}{c|c|ccccccccc}
\toprule
\textbf{Dataset} & \textbf{Metrics} & \textbf{self-MM} & \textbf{cube-MLP} & \textbf{DMD} & \textbf{DLF} & \textbf{TFRNet} & \textbf{MPLMM} & \textbf{MFMB-Net} & \textbf{LNLN} & \textbf{Ours}\\
\midrule
\multirow{6}{*}{\rotatebox[origin=c]{270}{CMU-MOSI}} & Acc-2 & 65.55 & \underline{69.61} & 69.46 & 68.80 & 50.31 & 59.20 & 69.14 & 67.79 & \textbf{75.15} \\
 & Acc-5 & \underline{35.37} & 34.70 & 29.64 & 34.21 & 17.35 & 22.35 & 32.59 & 33.14 & \textbf{40.82} \\
 & Acc-7 & \underline{31.78} & 30.91 & 26.78 & 29.93 & 17.20 & 21.77 & 29.07 & 30.03 & \textbf{33.09} \\
 & F1 & 65.00 & \underline{69.71} & 69.50 & 67.97 & 42.17 & 55.02 & 68.72 & 67.32 & \textbf{75.21} \\
 & MAE & \underline{1.071} & 1.106 & 1.096 & 1.084 & 1.425 & 1.325 & 1.082 & \textbf{1.066} & 1.165 \\
 & Corr & 0.556 & 0.542 & \underline{0.579} & \textbf{0.591} & 0.340 & 0.319 & 0.571 & 0.578 & 0.575 \\
\midrule
\multirow{6}{*}{\rotatebox[origin=c]{270}{SIMS-V2}} & Acc-2 & 63.73 & 65.25 & 68.60 & 71.02 & 61.77 & 62.99 & 71.12 & \underline{71.50} & \textbf{78.34} \\
 & Acc-3 & 48.48 & 44.13 & 52.90 & 58.61 & 45.36 & 37.56 & 60.47 & \underline{61.12} & \textbf{72.40} \\
 & Acc-5 & 33.33 & 30.72 & 36.52 & 39.46 & 31.08 & 25.73 & 40.57 & \underline{41.01} & \textbf{51.71} \\
 & F1 & 62.65 & 60.05 & 68.39 & 70.79 & 61.61 & 59.64 & \underline{70.98} & 70.59 & \textbf{78.36} \\
 & MAE & 0.432 & 0.452 & 0.416 & 0.413 & 0.490 & 0.477 & \underline{0.403} & 0.418 & \textbf{0.366} \\
 & Corr & 0.427 & 0.448 & 0.479 & 0.514 & 0.251 & 0.286 & \underline{0.514} & 0.513 & \textbf{0.610} \\
\midrule
\multirow{6}{*}{\rotatebox[origin=c]{270}{CMU-MOSEI}} & Acc-2 & 74.02 & 70.88 & 74.51 & \underline{75.42} & 71.71 & 70.06 & 74.20 & 75.19 & \textbf{77.27} \\
 & Acc-5 & 48.44 & 45.17 & 48.39 & 48.74 & 45.01 & 41.55 & \underline{48.92} & 48.79 & \textbf{49.15} \\
 & Acc-7 & 47.45 & 44.46 & 47.71 & 47.76 & 44.47 & 41.53 & \textbf{48.01} & \underline{47.79} & 47.75 \\
 & F1 & 72.21 & 65.10 & 73.70 & \underline{73.90} & 66.74 & 69.36 & 71.94 & 73.51 & \textbf{76.79} \\
 & MAE & 0.672 & 0.733 & 0.665 & \textbf{0.660} & 0.729 & 0.784 & 0.671 & 0.666 & \underline{0.660} \\
 & Corr & 0.568 & 0.572 & 0.587 & \underline{0.595} & 0.511 & 0.382 & 0.568 & 0.592 & \textbf{0.600} \\
\bottomrule
\end{tabular}
\caption {The performance of different methods on three datasets under Task 3. The best and second-best results are marked in bold and underlined, respectively.}
\label{tab:multi_dataset_results_3}
\end{table*}

\begin{table*}[t]
\centering
\setlength{\tabcolsep}{6pt}
\begin{tabular}{c|c|ccccccccc}
\toprule
\textbf{Dataset} & \textbf{Metrics} & \textbf{self-MM} & \textbf{cube-MLP} & \textbf{DMD} & \textbf{DLF} & \textbf{TFRNet} & \textbf{MPLMM} & \textbf{MFMB-Net} & \textbf{LNLN} & \textbf{Ours}\\
\midrule
\multirow{6}{*}{\rotatebox[origin=c]{270}{CMU-MOSI}} & Acc-2 & 68.70 & 68.50 & 67.53 & \underline{69.77} & 55.44 & 58.64 & 69.28 & 68.45 & \textbf{76.88} \\
 & Acc-5 & 33.58 & 33.04 & 31.68 & \underline{35.47} & 18.95 & 23.08 & 34.03 & 35.03 & \textbf{42.03} \\
 & Acc-7 & 29.98 & 29.50 & 28.47 & 30.56 & 17.64 & 22.30 & 30.21 & \underline{31.05} & \textbf{34.55} \\
 & F1 & 67.33 & 68.61 & 67.05 & \underline{69.89} & 49.83 & 55.80 & 68.12 & 68.14 & \textbf{76.80} \\
 & MAE & 1.062 & 1.099 & 1.093 & \underline{1.059} & 1.446 & 1.329 & 1.085 & \textbf{1.042} & 1.079 \\
 & Corr & 0.584 & 0.547 & 0.573 & \underline{0.592} & 0.326 & 0.332 & 0.566 & \textbf{0.592} & 0.590 \\
\midrule
\multirow{6}{*}{\rotatebox[origin=c]{270}{SIMS-V2}} & Acc-2 & 66.92 & 65.12 & 67.89 & 70.86 & 64.80 & 64.02 & 69.70 & \underline{71.47} & \textbf{77.69} \\
 & Acc-3 & 48.55 & 44.20 & 57.35 & 59.57 & 52.64 & 55.06 & 57.62 & \underline{60.67} & \textbf{70.54} \\
 & Acc-5 & 32.98 & 30.01 & 39.52 & 39.68 & 35.56 & 35.78 & 38.53 & \underline{40.30} & \textbf{50.42} \\
 & F1 & 64.41 & 59.95 & 66.77 & 70.48 & 64.24 & 63.34 & 69.08 & \underline{70.90} & \textbf{77.75} \\
 & MAE & 0.441 & 0.467 & 0.424 & 0.421 & 0.456 & 0.477 & \underline{0.416} & 0.419 & \textbf{0.363} \\
 & Corr & 0.423 & 0.432 & 0.457 & 0.498 & 0.382 & 0.319 & 0.487 & \underline{0.504} & \textbf{0.593} \\
\midrule
\multirow{6}{*}{\rotatebox[origin=c]{270}{CMU-MOSEI}} & Acc-2 & 74.42 & 74.38 & \underline{75.60} & 74.98 & 73.09 & 73.17 & 73.78 & 75.22 & \textbf{76.28} \\
 & Acc-5 & 47.98 & 46.51 & 47.23 & \underline{48.62} & 45.66 & 41.21 & 47.85 & 47.63 & \textbf{49.68} \\
 & Acc-7 & 47.14 & 45.57 & 46.61 & \underline{47.68} & 45.12 & 41.16 & 46.96 & 46.59 & \textbf{47.94} \\
 & F1 & 72.60 & 72.86 & \underline{73.77} & 73.64 & 71.15 & 71.76 & 70.34 & 73.27 & \textbf{75.86} \\
 & MAE & 0.675 & 0.693 & \underline{0.669} & \textbf{0.668} & 0.716 & 0.785 & 0.677 & 0.674 & 0.671 \\
 & Corr & 0.570 & 0.549 & 0.585 & 0.580 & 0.511 & 0.395 & 0.571 & \textbf{0.594} & \underline{0.592} \\
\bottomrule
\end{tabular}
\caption {The performance of different methods on three datasets under Task 4. The best and second-best results are marked in bold and underlined, respectively.}
\label{tab:multi_dataset_results_4}
\end{table*}

\begin{table*}[t]
\caption {The performance of different methods on three datasets under Task 5. The best and second-best results are marked in bold and underlined, respectively.}
\centering
\setlength{\tabcolsep}{6pt}
\begin{tabular}{c|c|ccccccccc}
\toprule
\textbf{Dataset} & \textbf{Metrics} & \textbf{self-MM} & \textbf{cube-MLP} & \textbf{DMD} & \textbf{DLF} & \textbf{TFRNet} & \textbf{MPLMM} & \textbf{MFMB-Net} & \textbf{LNLN} & \textbf{Ours}\\
\midrule
\multirow{6}{*}{\rotatebox[origin=c]{270}{CMU-MOSI}} & Acc-2 & 66.82 & \underline{70.17} & 69.87 & 70.02 & 53.86 & 56.40 & 68.03 & 67.63 & \textbf{76.67} \\
 & Acc-5 & \underline{34.50} & 33.77 & 32.51 & 32.31 & 20.12 & 20.02 & 32.19 & 30.22 & \textbf{41.98} \\
 & Acc-7 & \underline{30.90} & 29.69 & 28.48 & 28.67 & 19.29 & 19.78 & 28.52 & 27.70 & \textbf{35.28} \\
 & F1 & 66.23 & \underline{70.27} & 69.96 & 69.66 & 50.73 & 50.50 & 67.43 & 67.04 & \textbf{76.79} \\
 & MAE & \underline{1.072} & 1.078 & 1.076 & \textbf{1.065} & 1.420 & 1.341 & 1.104 & 1.111 & 1.120 \\
 & Corr & 0.568 & 0.573 & 0.587 & \underline{0.589} & 0.337 & 0.311 & 0.556 & 0.579 & \textbf{0.597} \\
\midrule
\multirow{6}{*}{\rotatebox[origin=c]{270}{SIMS-V2}} & Acc-2 & 64.22 & 64.57 & 68.67 & 68.47 & 64.22 & 65.89 & 66.96 & \underline{71.47} & \textbf{78.21} \\
 & Acc-3 & 43.00 & \underline{57.03} & 50.29 & 56.58 & 47.13 & 50.87 & 55.78 & 53.64 & \textbf{71.15} \\
 & Acc-5 & 30.50 & 38.39 & 35.11 & \underline{39.88} & 32.59 & 32.95 & 38.08 & 36.91 & \textbf{49.90} \\
 & F1 & 62.77 & 59.40 & 68.25 & 67.43 & 60.44 & 63.80 & 66.85 & \underline{70.94} & \textbf{78.20} \\
 & MAE & 0.428 & 0.471 & 0.416 & \underline{0.412} & 0.464 & 0.477 & 0.430 & 0.414 & \textbf{0.361} \\
 & Corr & 0.438 & 0.412 & 0.475 & 0.481 & 0.370 & 0.324 & 0.455 & \underline{0.501} & \textbf{0.612} \\
\midrule
\multirow{6}{*}{\rotatebox[origin=c]{270}{CMU-MOSEI}} & Acc-2 & 75.06 & 66.39 & 75.08 & 75.30 & 70.06 & 69.99 & 74.11 & \underline{75.35} & \textbf{76.94} \\
 & Acc-5 & \underline{48.82} & 45.48 & 48.03 & 48.42 & 46.10 & 41.56 & 48.52 & 48.37 & \textbf{49.40} \\
 & Acc-7 & \underline{47.73} & 44.80 & 47.25 & 47.47 & 45.40 & 41.42 & 47.46 & 47.34 & \textbf{48.07} \\
 & F1 & 72.79 & 61.60 & 73.38 & \underline{73.95} & 67.15 & 69.99 & 71.25 & 73.46 & \textbf{76.29} \\
 & MAE & 0.670 & 0.731 & 0.671 & \underline{0.667} & 0.712 & 0.779 & 0.674 & 0.670 & \textbf{0.658} \\
 & Corr & 0.568 & 0.571 & 0.579 & 0.579 & 0.515 & 0.399 & 0.562 & \textbf{0.593} & \underline{0.592} \\
\bottomrule
\end{tabular}
\caption {The performance of different methods on three datasets under Task 5. The best and second-best results are marked in bold and underlined, respectively.}
\label{tab:multi_dataset_results_5}
\end{table*}

\begin{table*}[t]
\centering
\setlength{\tabcolsep}{6pt}
\begin{tabular}{c|c|ccccccccc}
\toprule
\textbf{Dataset} & \textbf{Metrics} & \textbf{self-MM} & \textbf{cube-MLP} & \textbf{DMD} & \textbf{DLF} & \textbf{TFRNet} & \textbf{MPLMM} & \textbf{MFMB-Net} & \textbf{LNLN} & \textbf{Ours}\\
\midrule
\multirow{6}{*}{\rotatebox[origin=c]{270}{CMU-MOSI}} & Acc-2 & 64.43 & 66.77 & 67.83 & 65.85 & 49.95 & 65.55 & \underline{67.96} & 67.02 & \textbf{73.63} \\
 & Acc-5 & 31.58 & 27.36 & 27.45 & 26.04 & 17.54 & 20.80 & 30.66 & \underline{31.83} & \textbf{38.87} \\
 & Acc-7 & 28.67 & 25.12 & 25.80 & 24.30 & 17.49 & 20.26 & 27.33 & \underline{29.49} & \textbf{31.78} \\
 & F1 & 63.17 & 66.42 & \underline{67.94} & 65.58 & 44.55 & 64.53 & 67.79 & 66.46 & \textbf{73.63} \\
 & MAE & 1.137 & 1.191 & 1.170 & 1.171 & 1.460 & 1.342 & \underline{1.126} & \textbf{1.104} & 1.182 \\
 & Corr & 0.531 & 0.498 & 0.508 & 0.548 & 0.107 & 0.350 & \underline{0.553} & \textbf{0.566} & 0.552 \\
\midrule
\multirow{6}{*}{\rotatebox[origin=c]{270}{SIMS-V2}} & Acc-2 & 64.18 & 63.06 & \underline{71.24} & 70.18 & 60.80 & 61.77 & 70.80 & 70.86 & \textbf{76.72} \\
 & Acc-3 & 48.39 & 41.20 & 57.99 & 59.16 & 46.32 & 55.74 & 59.33 & \underline{60.38} & \textbf{69.70} \\
 & Acc-5 & 33.33 & 29.17 & 40.39 & \underline{41.39} & 30.14 & 36.04 & 40.47 & 39.56 & \textbf{49.68} \\
 & F1 & 62.89 & 57.91 & \underline{70.73} & 69.84 & 55.45 & 60.75 & 70.46 & 70.40 & \textbf{76.70} \\
 & MAE & 0.434 & 0.475 & 0.407 & \underline{0.405} & 0.515 & 0.480 & 0.408 & 0.421 & \textbf{0.384} \\
 & Corr & 0.424 & 0.361 & 0.501 & 0.501 & 0.282 & 0.353 & 0.500 & \underline{0.510} & \textbf{0.571} \\
\midrule
\multirow{6}{*}{\rotatebox[origin=c]{270}{CMU-MOSEI}} & Acc-2 & 73.78 & 67.19 & 74.24 & \underline{74.99} & 68.35 & 71.42 & 72.36 & 74.72 & \textbf{75.94} \\
 & Acc-5 & 46.89 & 44.05 & 47.88 & 47.30 & 43.49 & 41.59 & \textbf{47.97} & \underline{47.89} & 46.58 \\
 & Acc-7 & 46.22 & 43.46 & \underline{47.14} & 46.44 & 42.81 & 41.48 & \textbf{47.17} & 46.87 & 45.42 \\
 & F1 & 70.83 & 62.74 & 73.59 & \underline{73.96} & 60.86 & 70.39 & 68.76 & 73.14 & \textbf{74.92} \\
 & MAE & 0.684 & 0.751 & \textbf{0.674} & 0.682 & 0.781 & 0.773 & 0.680 & \underline{0.678} & 0.710 \\
 & Corr & 0.547 & 0.528 & \underline{0.575} & 0.567 & 0.477 & 0.390 & 0.552 & \textbf{0.583} & 0.534 \\
\bottomrule
\end{tabular}
\caption {The performance of different methods on three datasets under Task 6. The best and second-best results are marked in bold and underlined, respectively.}
\label{tab:multi_dataset_results_6}
\end{table*}

\begin{table*}[t]
\centering
\setlength{\tabcolsep}{6pt}
\begin{tabular}{c|c|ccccccccc}
\toprule
\textbf{Dataset} & \textbf{Metrics} & \textbf{self-MM} & \textbf{cube-MLP} & \textbf{DMD} & \textbf{DLF} & \textbf{TFRNet} & \textbf{MPLMM} & \textbf{MFMB-Net} & \textbf{LNLN} & \textbf{Ours}\\
\midrule
\multirow{6}{*}{\rotatebox[origin=c]{270}{CMU-MOSI}} & Acc-2 & 67.73 & \underline{70.38} & 67.53 & 69.26 & 66.21 & 61.74 & 68.89 & 67.07 & \textbf{77.03} \\
 & Acc-5 & 35.86 & \underline{36.44} & 32.95 & 36.00 & 24.73 & 21.53 & 34.92 & 34.45 & \textbf{43.29} \\
 & Acc-7 & 32.27 & \underline{33.04} & 29.50 & 31.97 & 21.28 & 20.89 & 31.10 & 30.95 & \textbf{36.64} \\
 & F1 & 67.06 & \underline{70.43} & 67.12 & 69.13 & 65.74 & 60.78 & 68.39 & 66.37 & \textbf{77.15} \\
 & MAE & \textbf{1.036} & 1.051 & 1.065 & \underline{1.038} & 1.326 & 1.365 & 1.053 & 1.058 & 1.101 \\
 & Corr & 0.595 & 0.576 & 0.590 & \underline{0.602} & 0.454 & 0.331 & 0.597 & 0.600 & \textbf{0.602} \\
\midrule
\multirow{6}{*}{\rotatebox[origin=c]{270}{SIMS-V2}} & Acc-2 & 66.44 & 62.89 & 68.38 & 69.70 & 63.60 & 66.60 & 69.93 & \underline{71.28} & \textbf{78.47} \\
 & Acc-3 & 42.91 & 42.30 & 57.99 & 59.57 & 51.71 & 56.77 & \underline{61.40} & 57.06 & \textbf{72.28} \\
 & Acc-5 & 30.72 & 28.63 & 39.43 & 41.91 & 35.94 & 37.52 & \underline{42.51} & 37.91 & \textbf{51.90} \\
 & F1 & 64.43 & 57.81 & 67.36 & 68.79 & 62.34 & 65.93 & 69.69 & \underline{70.23} & \textbf{78.57} \\
 & MAE & 0.431 & 0.480 & 0.421 & 0.412 & 0.446 & 0.455 & \underline{0.411} & 0.423 & \textbf{0.364} \\
 & Corr & 0.424 & 0.398 & 0.475 & 0.493 & 0.391 & 0.385 & 0.493 & \underline{0.505} & \textbf{0.608} \\
\midrule
\multirow{6}{*}{\rotatebox[origin=c]{270}{CMU-MOSEI}} & Acc-2 & 74.78 & 72.97 & 74.41 & \underline{75.71} & 73.28 & 73.03 & 73.47 & 75.31 & \textbf{75.97} \\
 & Acc-5 & 48.01 & 48.53 & \underline{49.13} & 48.95 & 47.60 & 42.51 & 48.65 & 49.00 & \textbf{49.40} \\
 & Acc-7 & 47.15 & 47.46 & \textbf{48.12} & 48.00 & 46.14 & 42.47 & 47.69 & \underline{48.05} & 47.96 \\
 & F1 & 71.86 & 71.61 & 73.03 & \underline{74.01} & 69.52 & 71.79 & 69.95 & 73.24 & \textbf{75.58} \\
 & MAE & 0.668 & 0.664 & \underline{0.662} & \textbf{0.659} & 0.704 & 0.775 & 0.670 & 0.663 & 0.662 \\
 & Corr & 0.574 & 0.583 & 0.590 & 0.592 & 0.541 & 0.415 & 0.575 & \underline{0.593} & \textbf{0.595} \\
\bottomrule
\end{tabular}
\caption {The performance of different methods on three datasets under Task 7. The best and second-best results are marked in bold and underlined, respectively.}
\label{tab:multi_dataset_results_7}
\end{table*}

\end{document}